\pdfoutput=1

\documentclass[11pt]{article}

\usepackage[final]{acl}
\usepackage{latexsym}
\usepackage{wrapfig}

\usepackage{amsmath,amsfonts,bm}

\def\eqref#1{equation~\ref{#1}}

\def\1{\bm{1}}

\DeclareMathAlphabet{\mathsfit}{\encodingdefault}{\sfdefault}{m}{sl}
\SetMathAlphabet{\mathsfit}{bold}{\encodingdefault}{\sfdefault}{bx}{n}

\usepackage{times}
\usepackage{color}
\usepackage{tabularx}
\usepackage{tabularray}
\usepackage{cleveref}
\crefname{section}{§}{§§}
\Crefname{section}{§}{§§}
\crefname{appendix}{§}{§§}
\Crefname{appendix}{§}{§§}
\crefname{equation}{Eq.}{Eqs.}
\Crefname{equation}{Equation}{Equations}
\usepackage[T1]{fontenc}

\usepackage[utf8]{inputenc}

\usepackage{microtype}

\usepackage{inconsolata}

\usepackage{graphicx}
\usepackage{microtype}

\usepackage{hyperref}       %
\usepackage{inconsolata}
\usepackage{wrapfig}
\usepackage{colortbl}
\usepackage{wrapfig}
\usepackage{xcolor}
\usepackage{adjustbox} 
\usepackage{inconsolata}
\usepackage{multirow}  
\usepackage{amssymb}   
\usepackage{amsmath}
\usepackage{color}
\usepackage{fontawesome5}
\usepackage{subfigure} 
\usepackage{makecell} 
\usepackage{booktabs} 
\usepackage{enumitem} 

\usepackage{stfloats}
\usepackage{listings}
\usepackage{cite}
\usepackage{pifont}
\usepackage{tcolorbox}
\usepackage{enumitem}
\tcbuselibrary{skins, breakable, theorems}
\usepackage{wrapfig}
\usepackage{lscape}
\usepackage{url}

\usepackage{siunitx}    
\usepackage[sectionbib]{chapterbib}
\usepackage{arydshln}
\usepackage{caption}         
\usepackage{array}           
\usepackage[table]{xcolor}   
\usepackage{varwidth}

\newcommand{\MDA}[0]{MDA\xspace}
\newcommand{\IFG}[0]{SIG\xspace}
\newcommand{\OURS}[0]{LACING\xspace}

\definecolor{Gray}{gray}{0.9}
\definecolor{mygreen}{rgb}{0.0, 0.5, 0.0}
\definecolor{myred}{rgb}{0.8, 0.25, 0.33}
\definecolor{myblue}{rgb}{0.19, 0.55, 0.91}
\definecolor{uclablue}{rgb}{0.15, 0.45, 0.68}
\definecolor{ucladblue}{rgb}{0.0, 0.33, 0.53}
\definecolor{ucladdblue}{rgb}{0.0, 0.23, 0.36}
\definecolor{uclagold}{rgb}{1.0, 0.82, 0.0}
\definecolor{ucladgold}{rgb}{1.0, 0.78, 0.17}
\definecolor{ucladdgold}{rgb}{1.0, 0.72, 0.11}
\definecolor{boxgreen}{rgb}{0.02, 0.66, 0.02}
\definecolor{boxred}{rgb}{0.66, 0.1, 0.1}
\definecolor{boxblue}{rgb}{0.01, 0.01, 0.73}

\usepackage{times}
\usepackage{latexsym}

\usepackage[utf8]{inputenc}
\usepackage[T1]{fontenc}

\usepackage{inconsolata}

\usepackage{algorithm}
\usepackage{algorithmic}

\usepackage{newfloat}
\usepackage{listings}
\floatstyle{ruled}
\newfloat{listing}{tb}{lst}{}
\floatname{listing}{Listing}

\usepackage{graphicx}
\usepackage{amsmath,amssymb,amsthm,mathabx,amsfonts}
\usepackage{bbm}
\usepackage{cleveref}
\usepackage{acronym}
\usepackage{enumitem}
\usepackage{balance}
\usepackage{xspace}
\usepackage{setspace}
\usepackage{url}
\usepackage{amsfonts}
\usepackage{multirow}
\usepackage{booktabs}
\usepackage{tablefootnote}
\usepackage[switch]{lineno}
\usepackage{multicol,lipsum}
\usepackage{tikz}
\usepackage{pgfplots}
\usepackage{pgfplotstable}
\usepackage{arydshln}

\usepackage{CJKutf8}

\pgfplotsset{compat=1.18}
\makeatletter
\DeclareRobustCommand\onedot{\futurelet\@let@token\@onedot}
\def\@onedot{\ifx\@let@token.\else.\null\fi\xspace}

\makeatother

\makeatletter
\newcommand{\thickhline}{%
    \noalign {\ifnum 0=`}\fi \hrule height 1pt
    \futurelet \reserved@a \@xhline
}

\crefname{algorithm}{Alg.}{Algs.}
\Crefname{algocf}{Algorithm}{Algorithms}
\crefname{section}{Sec.}{Secs.}
\Crefname{section}{Section}{Sections}
\crefname{table}{Tab.}{Tabs.}
\Crefname{table}{Table}{Tables}
\crefname{figure}{Fig.}{Fig.}
\Crefname{figure}{Figure}{Figure}
\crefname{appendix}{Appendix}{Appendices}

\acrodef{nlp}[NLP]{natural language processing}
\acrodef{plm}[PLM]{Pre-trained Language Model}
\acrodef{sota}[SOTA]{state-of-the-art}
\acrodef{icl}[ICL]{In-Context Learning}
\acrodef{bbl}[BBL]{BIG-bench Lite}

\usepackage{colortbl}
\definecolor{gblue}{HTML}{4285F4}
\definecolor{gred}{HTML}{DB4437}
\definecolor{ggreen}{HTML}{0F9D58}

\definecolor{mygray}{gray}{.92}
\definecolor{emphypurple}{rgb}{0.302, 0.055, 0.659}

\definecolor{highlightgreen}{HTML}{009901}
\definecolor{highlightred}{HTML}{FD6864}

\title{Looking Beyond Text: Reducing Language bias in Large Vision-Language Models via Multimodal Dual-Attention and Soft-Image Guidance}

\author{
Haozhe Zhao\textsuperscript{$\heartsuit$}\thanks{Equal Contribution.},
Shuzheng Si\textsuperscript{$\diamondsuit$}\footnotemark[1],
Liang Chen\textsuperscript{$\spadesuit$},
Yichi Zhang\textsuperscript{$\spadesuit$} \\
\textbf{Maosong Sun}\textsuperscript{$\diamondsuit$},
\textbf{Baobao Chang}\textsuperscript{$\spadesuit$}\thanks{Corresponding Author.} ,
and \textbf{Minjia Zhang}\textsuperscript{$\heartsuit$}\\
 \small \textsuperscript{$\heartsuit$}University of Illinois Urbana-Champaign 
 \small \textsuperscript{$\spadesuit$}Peking University 
 \small \textsuperscript{$\diamondsuit$}Tsinghua University
}

\begin{document}
\maketitle
\renewcommand{\thefootnote}{\fnsymbol{footnote}}
\renewcommand{\thefootnote}{\arabic{footnote}}

\begin{abstract}
Large vision-language models (LVLMs) have achieved impressive results in vision-language tasks. 
However,
LVLMs suffer from hallucinations caused by language bias, which neglects images while over-relying on text. We identify two reasons for the bias:
1). Different training scales between the LLM pretraining and LVLM alignment stage.
2). The learned inference bias due to short-term dependency of text data. 
Therefore, we propose \OURS, designed to address such bias with  Mu\underline{\textbf{L}}timodal Du\underline{\textbf{A}}l-attention Me\underline{\textbf{C}}han\underline{\textbf{I}}sm (\MDA) a\underline{\textbf{N}}d Soft-Image \underline{\textbf{G}}uidance (\IFG).
Specifically, \MDA adopts a \textbf{parallel dual-attention mechanism} that constructs separate attention for visual and text inputs to enhance integration of visual inputs across model. \IFG uses a \textbf{learnable soft visual prompt} during training and inference to replace visual inputs, designed to compel LVLMs to prioritize text inputs during inference. 
Experiments across different model architectures and scales demonstrate that \OURS effectively debiases LVLMs from their language bias, enhancing visual comprehension and reducing hallucinations without additional resources.\footnote{~The model and code will be available at \url{https://lacing-lvlm.github.io/}. Email: haozhez6@illinois.edu}
\end{abstract}


\section{Introduction}
\label{sec:intro}
Large Language Models (LLMs)~\citep{openai2023gpt4,dubey2024llama3} represent a significant milestone in natural language processing~\citep{yang2024gpt4tools,chatgpt,gemini1}. By incorporating visual encoders into LLMs~\citep{LLAVA,Qwen-VL}, the development of Large Vision-Language Models (LVLMs)~\citep{gpt4o,gemini1} has been accelerated, enabling them to handle both visual and text inputs. This facilitates various applications using LVLMs such as autonomous driving~\citep{DriveGPT4}, image creation~\citep{labs2025flux1kontextflowmatching,wu2025qwenimagetechnicalreport,wu2025nepautoregressiveimageediting,chen2024sparkvisionlanguageintelligence2dimensional} and medical assistants~\citep{li2023llavamedtraininglargelanguageandvision}.

State-of-the-art LVLMs, despite their advanced capabilities in handling  both modalities, often produce erroneous or irrelevant responses to input images~\citep{MMSTAR,lan2024surveyhallucinationlargevisual}.
he main reason behind such hallucinations is referred to as language bias~\citep{MMICL}, i.e., models sometimes ``ignore'' visual inputs and generate text responses solely based on text inputs.
However, prior studies have not comprehensively explored the origins of such bias. 
We suggest that this bias may emerges for the following two reasons:

\begin{figure*}[htbp]
    \centering
    \includegraphics[width=1.0\textwidth]{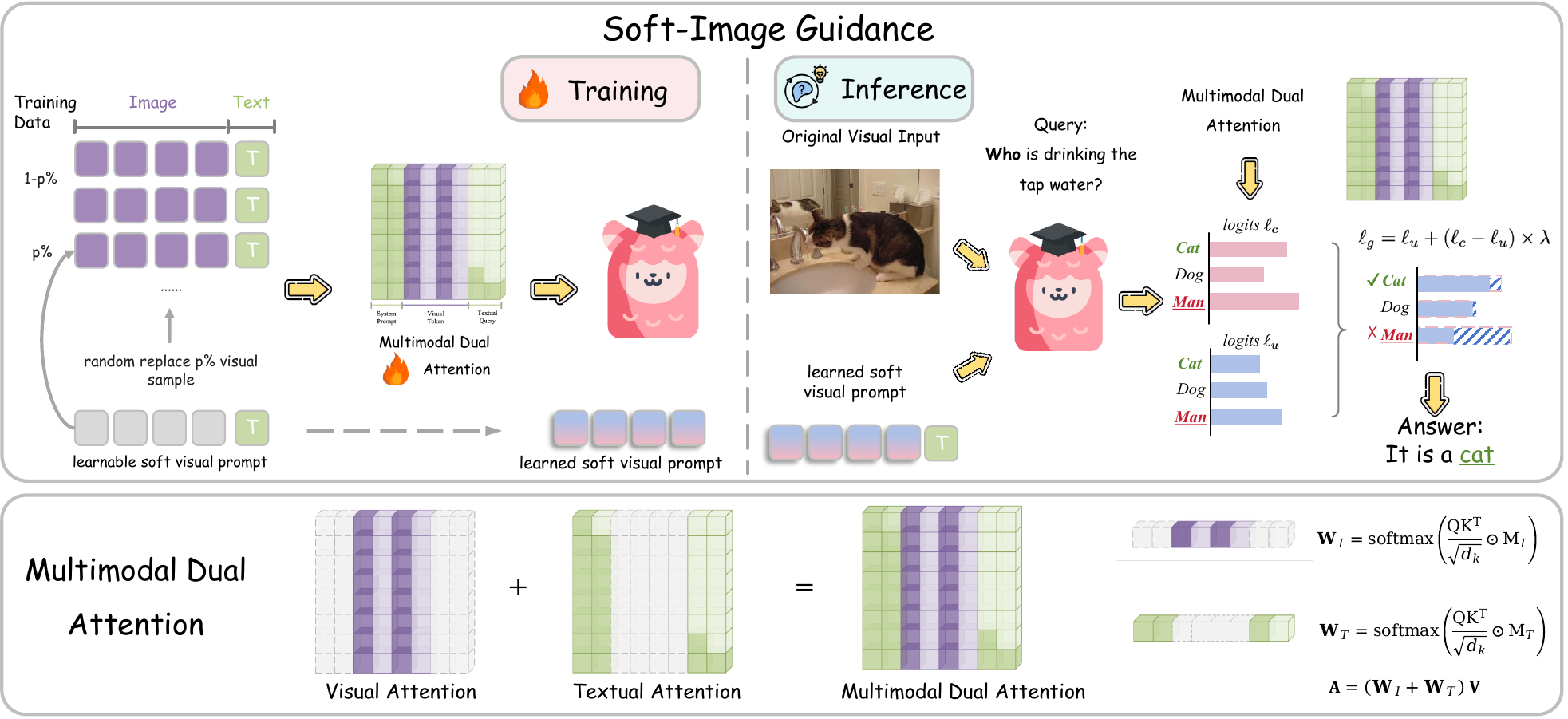} 
    \caption{Overview of \textbf{\OURS}, consisting of \textbf{Multimodal Dual Attention (bottom)} and \textbf{Soft-Image Guidance (above)} to mitigate language bias. \MDA proposes a \textbf{parallel} dual-attention mechanism that constructs two \textbf{separate} attention for visual and text inputs. \IFG implements a learnable soft visual prompt during training to replace visual inputs, which maintains input patterns while compelling model to prioritize text inputs during inference.
    }

    \label{fig:overview}
\end{figure*}

\textbf{1. Different training scales between pretraining and multimodal  alignment stage:}
The LLM backbone in LVLMs is pre-trained on on extensive text corpus,  while the multimodal alignment stage of LVLMs involves significantly fewer samples and shorter training duration.
For instance, Llama3~\citep{dubey2024llama3} is pre-trained with 15T tokens, whereas the multimodal alignment training for LLaVA-Series~\citep{LLAVA,LLAVA15,liu2024llavanext} employs only about 558k–1.3M examples.
This scale discrepancy causes the pretraining distribution to dominate the generation process in LVLMs~\citep{llavaBPO}, resulting in insufficient utilization of visual inputs.  As shown in \Cref{fig:avg_attention}, LVLMs allocate minimal attention to visual tokens in over 90\% layers~\citep{FastV}. Conversely, as discussed in \Cref{sec:early-fusion}, models such as Chameleon~\citep{Chameleon}, pretrained with balanced scales of textual and visual tokens, exhibit significantly reduced bias, further supporting this hypothesis.


\textbf{2. The learned inference bias due to the short-term dependency of text data}: Intuitively, a word in a text sequence exhibits a stronger associative bond with adjacent words than those further apart~\citep{alabdulmohsin2024fractalpatternsilluminatesuccess,daniluk2017frustratinglyshortattentionspans,yan2024recurformertransformerheadsneed}, i.e., the short-term dependency of text data. LLMs pre-trained on large-scale text corpora are more easily capturing and memorizing such short-term dependency~\citep{NSA}, which typically assign higher attention weights to adjacent tokens. However, this learned pattern may be problematic in multi-modal contexts. 
In current LVLMs, visual features are typically concatenated with text inputs to form input context. 
As generation progresses, the model increasingly focuses on nearby generated text tokens while progressively neglecting fixed-position visual inputs~\citep{zhang2024debiasingmultimodallargelanguage}, as shown in \Cref{fig:avg_attnntion_by_tokens}.



These two reasons lead to a systemic bias in LVLMs, originating from both training and inference stages. 
Consequently, a critical question arises: \textbf{\textit{How can we effectively mitigate language bias of LVLMs from both training and inference perspectives?}}  
Therefore, we propose \textbf{\OURS}, a systemic framework designed to address the language bias of LVLMs with Mu\underline{\textbf{L}}timodal Du\underline{\textbf{A}}l-attention Me\underline{\textbf{C}}han\underline{\textbf{I}}sm a\underline{\textbf{N}}d Soft-Image \underline{\textbf{G}}uidance.
To address training scale gaps in LVLMs, which leads to neglect of visual inputs across most layers~\citep{FastV}, we propose \textbf{Multimodal Dual-Attention Mechanism (\MDA)}.
Specifically, \MDA introduces a parallel dual-attention mechanism that separately computes attention weights for each modality, and then fuses them to form the final attention map.
This design ensures model to maintain substantial attention to visual inputs across all layers, promoting more effective visual-text integration.
Crucially, unlike previous methods that apply bidirectional attention to visual inputs within a shared attention matrix~\citep{Showo,Transfusion}, \MDA builds parallel attention map that compute modality-specific attention scores separately. This separation enables flexible attention configurations; for instance, visual inputs can adopt either causal or bidirectional attention.
In our design, we employ bidirectional attention for visual inputs to better capture global visual feature, while retaining causal attention for text to preserve the language modeling capabilities of LLMs. 
To mitigate learned inference bias in LVLMs, we propose \textbf{Soft-Image Guidance (\IFG)}, designed to enhance visual guidance by addressing the model’s over-reliance on textual inputs (i.e., language bias).
At core of \IFG is a \textbf{learnable soft visual prompt}, which replaces visual inputs during both training and inference.
It serves as a modality-aware placeholder, preserving input patterns (e.g., the input length and modalities), while implicitly compelling model to prioritize text inputs. 
Unlike prior methods~\citep{vcd,zhang2024debiasingmultimodallargelanguage} that remove visual inputs or inject random noise, \IFG maintains input consistency without introducing uncontrolled perturbations. 
During multimodal alignment stage, visual inputs are randomly replaced with soft prompt, allowing model to learn from complete and visual-substituted inputs.
At inference, we replace visual inputs with well-learned soft prompt to form multimodal-null input. 
Each token's final output is computed by contrasting model's output distributions from original and multimodal-null inputs, ensuring each token in responses accounts for visual input more critically and thereby reducing language bias.

Our proposed MDA and SIG form a systematic framework for mitigating language bias in LVLMs, with each component complementing the other to further enhance overall performance.
Comprehensive experiments across various model architectures and scales validate the effectiveness of \OURS. 
We observe significant improvements, particularly in free-form generation and visual hallucinations reduction (e.g., 11.8-point gain on LLaVA-Bench~\citep{LLAVA} and a 40\% improvement on Object Hall~\citep{objhal,RLHF-V}). 
Notably, \OURS delivers consistent improvement without additional resource requirements beyond standard multimodal alignment setups~\citep{LLAVA15,liu2024llavanext}.
Our analysis further confirms the efficacy of \MDA in enabling LVLMs to fully utilize visual inputs, and robustness of \IFG for reducing hallucinations and improving visual comprehension.





\section{Related Work}
\label{sec:related_work}

\subsection{Language Bias in LVLMs}
Despite the impressive capabilities of LVLMs~\citep{gpt4o,gemini1,MM1,wang2024qwen2vl,Llava-onevision,}, these models still struggle with generating responses irrelevant to the input images~\citep{lan2024surveyhallucinationlargevisual,liu2024surveyhallucinationlargevisionlanguage}, e.g., hallucinating non-existent objects~\citep{zhou2024analyzingmitigatingobjecthallucination}.
\citet{MMICL} first identify this issue in LVLMs and name it as \textit{language bias}, i.e., LVLMs often ignore visual inputs and solely rely on text inputs, leading to hallucinations.
\citet{MMSTAR} observe that LVLMs often answer questions using only LLM-derived textual knowledge.
\citet{FastV} further show that attention to visual inputs diminishes significantly in deeper layers, while \citet{zhang2024debiasingmultimodallargelanguage} find that models increasingly prioritize text as generation progresses.
These findings collectively indicate that LVLMs assign disproportionately low attention to visual inputs, limiting their ability to effectively utilize image information.
Therefore, to address this challenge, we propose a systematic framework, \OURS, that mitigates language bias from both training and inference perspectives.




\subsection{Addressing Language Bias in LVLMs} 

Given the language bias of LVLMs, they exhibit similar hallucination issues as LLMs~\citep{llmHall}, as well as modality-specific hallucinations such as object hallucination~\citep{objhal,pope}. 
As noted by~\citet{vcd}, this stems from the dominant influence of the LLM's pretraining distribution, making hallucination a prominent symptom of language bias.
Recent efforts to mitigate hallucination fall into two main categories. The first includes training-intensive methods such as LRV~\citep{lrv}, LLaVA-BPO~\citep{llavaBPO}, LLaVA-RLHF~\citep{LLAVA-RLHF}, and RLHF-V~\citep{RLHF-V}, which rely on supervised fine-tuning or reinforcement learning with preference data.
While effective, these methods typically necessitate substantial training data and computational resources. To address this, training-free methods have been proposed, including VCD~\citep{vcd}, IBD~\citep{IBD}, VDD~\citep{zhang2024debiasingmultimodallargelanguage}, and ICD~\citep{ICD}. These methods contrast outputs with those from image-free inputs (or with distorted images) to reduce influence of textual LLMs. However, these methods may introduce inconsistencies between training and inference, limiting their effectiveness.
Inspired by classifier-free guidance~\citep{ho2022classifierfreediffusionguidance}, which combines conditional and unconditional signals for image generation, we propose a novel approach that addresses language bias from both training and inference perspectives and targets broader bias effects beyond object hallucination, improving general LVLM performance.

\section{Method}
\label{sec:method}




\begin{figure}[t]  
  \centering  
    \includegraphics[width=\linewidth]{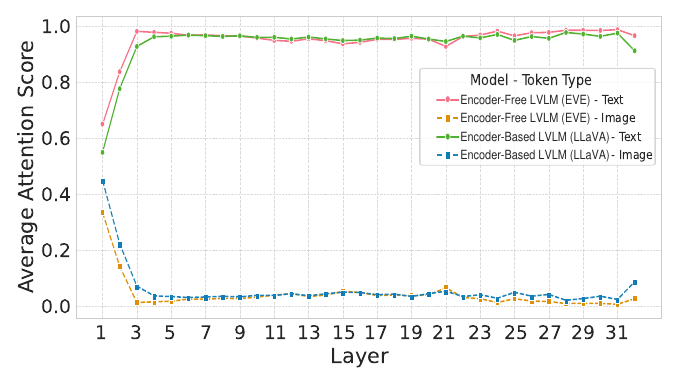}
  \caption{Average attention scores for output tokens towards text and visual tokens across different layers of encoder-based LVLMs~\citep{LLAVA15} and encoder-free LVLMs~\citep{diao2024EVE}, 
  showing that only the first two layers apply considerable attention to visual tokens. In contrast, deeper layers largely neglect them.}
  
  \label{fig:avg_attention}  
\end{figure}

\subsection{Multimodal Dual-Attention Mechanism}

Most LVLMs project bidirectional visual inputs into unidirectional LLM space using a relatively small amount of multimodal data~\citep{LLAVA,LLAVA15,Llava-onevision} compared to vast pretraining data scales of LLMs~\citep{dubey2024llama3}. 
LVLMs treat visual inputs as a different form of text inputs in an autoregressive manner. The mismatch in both modeling and training scale leads LVLMs to partially adapt to data distribution changes using only shallow layers during training with limited data~\citep{zhang2024debiasingmultimodallargelanguage}. Consequently, LVLMs remains dominated by LLM's pretraining distribution and lacks effective attention to visual inputs in deeper layers. Shown in \Cref{fig:avg_attention}, LVLMs~\citep{ Qwen-VL,wang2024qwen2vl,liu2024llavanext,diao2024EVE} exhibit considerable attention toward visual inputs only in the first two layers~\citep{FastV}, while deeper layers retain their original distributions, causing deeper layers of LVLMs to ignore visual inputs. 
This pheromone has been observed across \textbf{various LVLMs}, including encoder-based LVLMs, such as LLAVA-Series~\citep{LLAVA,LLAVA15,liu2024llavanext}, QwenVL~\citep{Qwen-VL} and Qwen2VL~\citep{wang2024qwen2vl}, and even encoder-free LVLMs like EVE~\citep{diao2024EVE} and Fuyu~\citep{fuyu}.

To address this issue, we propose \textbf{Multimodal Dual-Attention Mechanism (\MDA)}, which introduces a \textbf{parallel} dual-attention mechanism that preserves separate attention metrics for visual and text inputs in the LVLMs.
It enforces LLMs to allocate sufficient attention toward visual inputs and encourages LVLMs to fully leverage their LLM backbone for visual comprehension during training.
This separation enables flexible attention configurations; for instance, visual inputs can adopt either causal or bidirectional attention.
In our design, \MDA retains causal attention for text inputs while independently calculating bidirectional attention towards visual inputs.
As illustrated in Equation~\ref{eq:multimodal_attn}, given multimodal inputs $\mathbf{S} = \langle s_1, s_2, \dots, s_N \rangle$, 
$s_n$ means the token in inputs.
To independently calculate attention weights across two modalities, we define two attention masks: mask $\mathbf{M}_{\mathcal{I}}$ for visual tokens $\mathcal{I}$ and mask $\mathbf{M}_{\mathcal{T}}$ for text tokens $\mathcal{T}$:
\begin{equation}
\small 
\begin{aligned}
    \mathbf{M}_{\mathcal{I}}[i, j] &= 
    \begin{cases}
        1, & \text{if } s_j \in \mathcal{I}, \\
        0, & \text{otherwise},
    \end{cases} \\
    \mathbf{M}_{\mathcal{T}}[i, j] &= 
    \begin{cases}
        1, & \text{if }  s_j \in \mathcal{T} \And i \leq j, \\
        0, & \text{otherwise},
    \end{cases}
\end{aligned}
\label{eq:multimodal_attn}
\end{equation}


We use the attention masks to calculate attention weights of visual$(\mathbf{W}_{\mathcal{I}})$ and text tokens$(\mathbf{W}_{\mathcal{T}})$:

\begin{equation}
\small
\begin{aligned}
    \mathbf{W}_{\mathcal{I}} &= \text{softmax}\left(\mathbf{Q} \mathbf{K}^\top / \sqrt{d_k} \odot \mathbf{M}_{\mathcal{I}} \right), \\
    \mathbf{W}_{\mathcal{T}} &= \text{softmax}\left( \mathbf{Q} \mathbf{K}^\top / \sqrt{d_k} \odot \mathbf{M}_{\mathcal{T}} \right), \\
\end{aligned}
\label{eq:attn_Wcal}
\end{equation}

where $\mathbf{Q},\mathbf{K}$ is query, key and in self-attention of LVLMs.
Finally, the two attention weights $(\mathbf{W}_{\mathcal{I}})$ and $(\mathbf{W}_{\mathcal{T}})$, are integrated and multiplied by $\mathbf{V}$, the value in attention mechanism, to derive final attention score $\mathbf{A}$ based on \MDA.
\begin{equation}
\small
    \mathbf{A} = \left( \mathbf{W}_{\mathcal{I}} + \mathbf{W}_{\mathcal{T}} \right) \mathbf{V}.
\label{eq:attn}
\end{equation}

\begin{figure}[t]  
  \centering  
  \includegraphics[width=\linewidth]{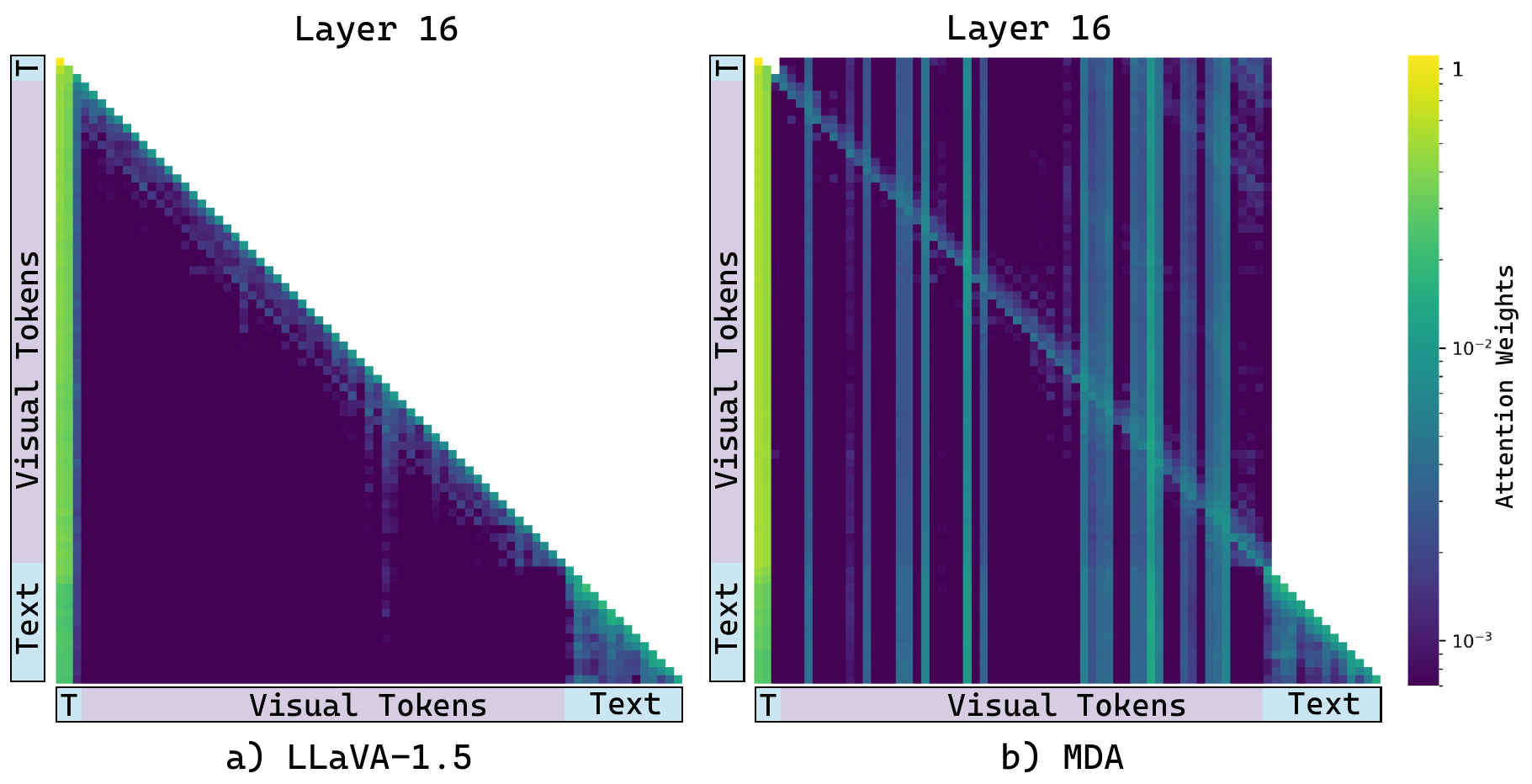}  
 \caption{ Attention allocation of a standard LVLM (LLaVA-1.5) and model trained with \MDA. Text and visual tokens are marked in \textbf{\textcolor[HTML]{a6cdd8}{blue}} and \textbf{\textcolor[rgb]{0.675, 0.631, 0.725}{purple}}, respectively.}

  \label{fig:compare}  
\end{figure}

Parallel computation of attention weights guarantees each token separately receives attention from both visual and text inputs, balancing their contributions. It allows visual inputs to remain relevance across all layers, avoiding shallow adaptation and language bias. 
\MDA ensures that visual information is processed with bidirectional attention to capture spatial coherence, while text tokens continue to follow autoregressive patterns, critical for maintaining coherent language generation, as shown in \Cref{fig:compare}. To support this design choice, we present a comparison between causal and bidirectional attention for visual inputs in \Cref{sec:design_choice}.

\subsection{Soft-Image Guidance}
\begin{figure}[t]  
  \centering  
  \includegraphics[width=\linewidth]{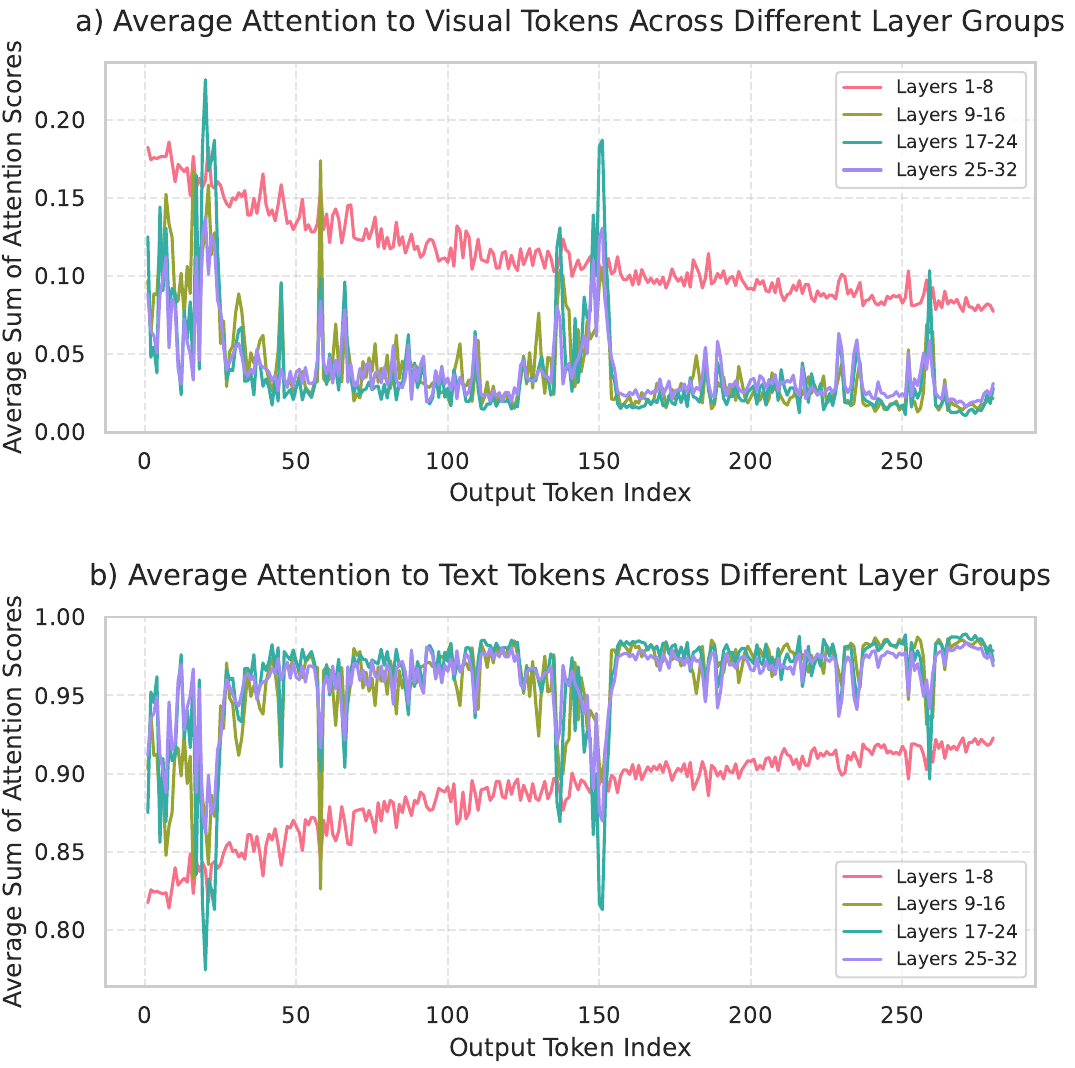}  
  \caption{Attention allocation to visual and text tokens. Attention to visual tokens (a) decreases as response generates, while attention to text tokens (b) increases.}  
  \label{fig:avg_attnntion_by_tokens}  
\end{figure}

Due to the sequential nature of language modeling, which prioritizes coherence and continuity, LVLMs tend to focus on nearby text tokens, often at the expense of the visual information that may be distant or disparate, as shown in \Cref{fig:avg_attnntion_by_tokens}.


Inspired by classifier-free guidance~\citep{ho2022classifierfreediffusionguidance} effectively combining the conditional and unconditional score to control the image generation quality, we propose the Soft-Image Guidance (\IFG), designed to enhance the guidance of visual inputs during LVLMs’ response generation and mitigate the inference bias of LVLMs.
To enhance the guidance of visual inputs in LVLMs, we formulate the visual comprehension mathematically. We consider the conditional probability $p(y_t\mid v)$ of generating a response token \( y_t \) given the visual input \( v \). By applying Bayes' theorem, we have:  
\begin{equation}  
\small
\label{eq:conditional_generation}  
p(y_t \mid v) = \frac{p(v \mid y_t) \cdot p(y_t)}{p(v)}  
\end{equation}  
Then we take the logarithm of both sides of \cref{eq:conditional_generation}: 
\begin{equation}  
\small
\label{eq:log_conditional_generation}  
\log p(y_t \mid v) = \log p(v \mid y_t) + \log p(y_t) - \log p(v)  
\end{equation}  
In \cref{eq:log_conditional_generation}, \( p(y_t) \) is unconditional probability of generating token \( y_t \) without visual input.  

To amplify influence of visual input \( v \) on text generation, we introduce a scaling parameter \( \lambda \) for conditional probability \( p(v \mid y_t) \). We adjust  \( p(v \mid y_t) \) to obtain an enhanced version \( \widehat{p}(y_t \mid v) \):  
\begin{equation}  
\small
\label{eq:scaled_log_likelihood}  
\log \widehat{p}(y_t \mid v) \propto \lambda \cdot \log p(v \mid y_t) + \log p(y_t) - \log p(v)  
\end{equation}  
To express \( \log \widehat{p}(y_t \mid v) \) with known quantities, we expand \( \log p(v \mid y_t) \) using Bayes' theorem:  
\begin{equation}  
\small
\label{eq:log_p_v_given_yt}  
\log p(v \mid y_t) = \log p(y_t \mid v) + \log p(v) - \log p(y_t)  
\end{equation}  

Substituting \cref{eq:log_p_v_given_yt} into \cref{eq:scaled_log_likelihood}, we obtain:  
\begin{equation} 
\small
\label{eq:expanded_log_likelihood}  
\begin{aligned}  
\log \widehat{p}(y_t \mid v) &\propto \lambda \left( \log p(y_t \mid v) + \log p(v) - \log p(y_t) \right)  \\
&\quad + \log p(y_t) - \log p(v).  
\end{aligned}  
\end{equation}
Since \( v \) is given (fixed), \(\log p(v)\) is constant for \(y_t\)and can be omitted, we simplify \cref{eq:expanded_log_likelihood} to:  
\begin{equation}  
\small
\label{eq:simplified_log_likelihood}  
\log \widehat{p}(y_t \mid v) \propto \lambda (\log p(y_t \mid v) - \log p(y_t) ) + \log p(y_t)
\end{equation} 

\cref{eq:simplified_log_likelihood} demonstrates that influence of visual input \( v \) on text generation can be amplified by adjusting scaling parameter \( \lambda \), once given conditional probability \( p(y_t \mid v) \) of original inputs and unconditional probability \( p(y_t) \) without visual inputs. This formulation highlights a major challenge in enhancing visual guidance for LVLMs: accurately calculating unconditional probability \( p(y_t) \) of generating token \( y_t \) in the absence of visual input.

\begin{algorithm}[t]  
    \caption{Joint Training of LVLM with \IFG}  \small
    \label{alg:image_free_guidance}  
    \begin{algorithmic}[1]  
        \REQUIRE $P$: Model; $\mathcal{X}, \mathcal{V}$: Training dataset 
        \REPEAT  
            \STATE $(\mathbf{x}, \mathbf{v}) \sim (\mathcal{X}, \mathcal{V})$ \hfill $\triangleright$ Sample multimodal input data
            \STATE $\mathbf{v} \gets \epsilon$ \textbf{with probability} $\theta$ \hfill $\triangleright$ Replace visual input with soft prompt $\epsilon$
            \STATE $\mathcal{L}_{\text{cross-entropy}} = -\mathbb{E}_{(\mathbf{x}, \mathbf{v})} \sum\limits_{i} \mathbf{y}_i \log P(\mathbf{x}, \mathbf{v})$
            \STATE Update $P$ and $\epsilon$
        \UNTIL{converged}  
    \end{algorithmic}  
\end{algorithm}

Previous approaches attempt to ascertain such probabilities probability by either providing the model with text-only input~\citep{zhang2024debiasingmultimodallargelanguage} or by injecting randomly generated noise to mask the image~\citep{vcd}, thereby utilizing the model’s output as the unconditional probability \( p(y_t) \). Nonetheless, simply removing the visual inputs may disrupt input patterns(e.g., the input length and modalities), as visual tokens typically far surpass text tokens in quantity~\citep{FastV,zhang2024debiasingmultimodallargelanguage}. Concurrently, adding random noise to distort images relies can introduce uncontrollable and unstable informational perturbations. The extra, unforeseen noise introduced by these inputs may lead the LVLMs to behave more like random probability generators, thereby complicating the approximation of \( p(y_t) \).

\begin{table*}[ht]  
  \centering  
  \scriptsize  
  \resizebox{0.87\textwidth}{!}{  
    \begin{tabular}{@{}lccccccc@{}}  
        \toprule  
      \multirow{2}{*}{Model} & \multirow{2}{*}{Model Size} & \multicolumn{2}{c}{Obj Hall} & \multicolumn{2}{c}{MMHall} & \multirow{2}{*}{LLaVABench$\uparrow$} & \multirow{2}{*}{MM-VET$\uparrow$} \\
      \cmidrule(lr){3-4} \cmidrule(lr){5-6}  
       &  & Res $\downarrow$ & Obj $\downarrow$ & Score $\uparrow$ & Hall $\downarrow$ &  &  \\
      \midrule  
      LRV$^\dag$~\citep{lrv}  & 7B & 32.30 & 22.30 & / & / & / & 31.70 \\
      LLaVA-1.5$^\dag$~\citep{LLAVA15} & 7B & 46.71 & 25.08 & 2.19 & 59.00 & 64.40 & 31.10 \\
      \hdashline[2pt/3pt]
      VCD$^\ddag$~\citep{vcd} & 7B & 47.40 & 25.24 & 2.12 & 59.00 & 65.30 & 30.90 \\
      VDD-None$^\ddag$~\citep{zhang2024debiasingmultimodallargelanguage} & 7B & 46.71 & 25.19 & 2.22 & 56.00 & 66.00 & 31.70 \\
      ICD$^\ddag$~\citep{ICD} & 7B & 47.40 & 25.00 & 2.18 & 59.00 & 64.70 & 31.10 \\
      Less-is-more$^\ddag$~\citep{Less} & 7B & 40.30 & 17.80 & \underline{2.33} & \underline{50.00} & 60.90 & / \\
      OPERA$^\ddag$~\citep{OPERA} & 7B & 45.10 & 22.30 & 2.15 & 54.20 & 60.30 & / \\
      \hdashline[2pt/3pt]
      HA-DPO$^\circ$~\citep{HA-DPO}  & 7B & 39.90 & 19.90 & 1.98 & 60.40 & 67.20 & / \\
      POVID$^\circ$~\citep{POVID}  & 7B & 48.10 & 24.40 & 2.08 & 56.20 & 62.20 & / \\
      LLaVA1.5-7B-BPO$^\circ$~\citep{llavaBPO} & 7B & \underline{31.90} & \underline{15.10} & / & / & \underline{71.60} & \textbf{36.80} \\
      \hdashline[2pt/3pt]
     \rowcolor{blue!5} \bf \OURS  & 7B & \textbf{27.86} & \textbf{14.22} & \textbf{2.53} & \textbf{49.00} & \textbf{72.20} & \underline{35.20} \\
     \rowcolor{blue!5} $\Delta$, compare to LLaVA-1.5 & 7B & 40.36\% & 43.30\% & 15.53\% & 16.95\% & 12.11\% & 13.18\% \\
      \midrule  
      LLaVA$^\dag$~\citep{LLAVA}  & 13B & 63.00 & 29.50 & / & / & 70.80 & 26.40 \\
      Muffin$^\dag$~\citep{MUFFIN} & 13B & 50.50 & 24.50 & / & / & 68.80 & / \\
      QWEN-VL$^\dag$~\citep{Qwen-VL}  & 10B & 40.40 & 20.70 & / & / & 52.10 & / \\
      LLaVA-1.5$^\dag$~\citep{LLAVA15} & 13B & 47.06 & 23.33 & 2.54 & 50.00 & 72.50 & 36.10 \\
      \hdashline[2pt/3pt]
      
      VCD$^\ddag$~\citep{vcd} & 13B & 46.37 & 23.10 & \underline{2.60} & \underline{49.00} & 73.60 & 36.90 \\
      VDD-None$^\ddag$~\citep{zhang2024debiasingmultimodallargelanguage} & 13B & 44.64 & 22.23 & 2.38 & 55.00 & 73.00 & 36.10 \\
      ICD$^\ddag$~\citep{ICD} & 13B & 45.52 & 21.93 & 2.41 & 54.00 & 72.50 & 36.20 \\
      \hdashline[2pt/3pt]
      
      LLaVA-RLHF$^\circ$~\citep{LLAVA-RLHF} & 13B & 38.10 & 18.90 & 2.53 & 57.00 & 61.50 & / \\
      RLHF-V$^\circ$~\citep{RLHF-V} &  13B & \textbf{12.20} & \textbf{7.50} & 2.45 & 51.00 & 51.40 & / \\
      LLaVA1.5-13B-BPO$^\circ$~\citep{llavaBPO} & 13B & 27.30 & \underline{12.90} & / & / & \underline{74.40} & \textbf{41.40} \\
      \hdashline[2pt/3pt]
     \rowcolor{blue!5} \bf \OURS  & 13B & \underline{27.21} & 14.10 & \textbf{2.65} & \textbf{48.00} & \textbf{84.30} & \underline{39.90} \\
      \rowcolor{blue!5} $\Delta$, compare to LLaVA-1.5 & 13B & 42.18\% & 39.56\% & 4.33\% & 4.00\% & 16.28\% & 10.53\% \\
      \bottomrule  
    \end{tabular}  
  }  
  \caption{Comparison across multiple benchmarks, highlighting highest score in \textbf{bold} and second highest \underline{underlined}. Baselines are categorized as: $\dag$ (LVLMs), $\ddag$ (training-free), and $\circ$ (reinforcement learning-based).}
  \label{tab:comparison}  
\end{table*}

\IFG first employs a learnable soft visual prompt \( \epsilon \) to replace the visual input, thereby forming a multimodal-null input for the model. The learnable soft visual prompt \( \epsilon \) will be the jointly trained with the LVLM. 
As outlined in Algorithm~\ref{alg:image_free_guidance}, we replace visual input with \( \epsilon \) with probability \( \theta \) during training. The soft visual prompt \( \epsilon \) serves a dual purpose, acting both as a placeholder to maintain the input pattern and as an indicator to make the model prioritize text input.
This dual functionality ensures a consistent input pattern for LVLMs in both training and inference, allowing the model to produce generate interpretable output and balancing the visual and text inputs.
After training, we can directly use the \( \epsilon \) to query the model and extract the approximation of \( p(y_t) \).
Finally, during inference, we contrast output distributions from original and multimodal-null inputs based on Equation~\ref{eq:simplified_log_likelihood}  to get the final output. Specifically, logits \( \ell_{g} \) of generated tokens are recalculated by adjusting the logits \( \ell_{u} \) of the multimodal-null inputs with the scaling parameter \( \lambda \), based on logits \( \ell_{c} \) of original inputs as follows:
\begin{equation}  
\small
\label{eq:cfg}
\ell_{g} = \ell_{u} + (\ell_{c} - \ell_{u}) \times \lambda  
\end{equation} 

\cref{eq:cfg} facilitates a more balanced and effective integration of visual inputs, enhancing visual comprehension while addressing the language bias.








\section{Experiments}


\subsection{Implementation Details}
To ensure fair comparison and validate the effectiveness of our approach, we train LVLMs from scratch and evaluate against strong baselines.
Given availability of open-sourced multimodal alignment datasets, we select two representative LVLMs with different architectures and model scales: LLaVA-1.5~\citep{LLAVA15} and LLaVA-Next~\citep{liu2024llavanext} as our base model. We strictly follow their training settings, including the same dataset and model backbone.
The model is trained on 8 A100 GPUs, each with 40 GB of memory. Details of scaling parameter $\lambda$ and replacement probability $\theta$ are shown in \Cref{sec:Hyperparameters}.
Additional information, including extra costs discussion, training and experiment details, can be found in \Cref{sec:Additional}, \Cref{sec:Training_Details}, \Cref{sec:Details}, and \Cref{sec:Experimental_detail}.

\begin{table*} [t]
  \centering  
  \scriptsize  

  \resizebox{0.8\textwidth}{!}{  
    \begin{tabular}{@{}lcccccc@{}}  
      \toprule  
      \multirow{2}{*}{Method} & \multirow{2}{*}{Model Size} & \multirow{2}{*}{MMBench$\uparrow$} & \multirow{2}{*}{TextVQA$\uparrow$} & \multirow{2}{*}{LLaVABench$\uparrow$} & \multicolumn{2}{c}{Obj Hall} \\   
      \cmidrule(lr){6-7}  
      & & & & & Res $\downarrow$ & Obj $\downarrow$ \\   
      \midrule  
      \multicolumn{7}{c}{\cellcolor{olive!10}\textbf{Greedy Sampling}} \\   
      \midrule  
      LLaVA-1.5 & 7B & 64.61 & 46.05 & 64.40 & 46.71 & 25.08 \\
      VCD & 7B & 64.69 (\textcolor[rgb]{0.7,0,0}{+ 0.08}) & 46.05 (\textcolor[rgb]{0,0.7,0}{+ 0.00}) & 65.30 (\textcolor[rgb]{0.7,0,0}{+ 0.90}) & 47.40 (\textcolor[rgb]{0,0.7,0}{+  0.69}) & 25.24 (\textcolor[rgb]{0,0.7,0}{+ 0.16}) \\
      VDD-None & 7B & 64.52 (\textcolor[rgb]{0,0.7,0}{- 0.09}) & 44.47 (\textcolor[rgb]{0,0.7,0}{- 1.58}) & 66.00 (\textcolor[rgb]{0.7,0,0}{+ 1.60}) & 46.71 (\textcolor[rgb]{0,0.7,0}{+  0.00}) & 25.19 (\textcolor[rgb]{0,0.7,0}{+ 0.10}) \\
      \rowcolor{blue!5} \textbf{w. \IFG} & 7B & \textbf{66.92} (\textcolor[rgb]{0.7,0,0}{+ 2.31}) & \textbf{46.77} (\textcolor[rgb]{0.7,0,0}{+ 0.72}) & \textbf{70.60} (\textcolor[rgb]{0.7,0,0}{+ 6.20}) & \textbf{30.36} (\textcolor[rgb]{0.7,0,0}{- 16.35}) & \textbf{15.16} (\textcolor[rgb]{0.7,0,0}{- 9.92}) \\
      \midrule  
      \multicolumn{7}{c}{\cellcolor{olive!10}\textbf{Nucleus Sampling}} \\   
      \midrule  
      LLaVA-1.5 & 7B & 56.96 & 35.41 & 63.00 & 56.66 & 29.75 \\
      VCD & 7B & 60.91 (\textcolor[rgb]{0.7,0,0}{+ 3.95}) & 40.67 (\textcolor[rgb]{0.7,0,0}{+ 5.26}) & 65.30 (\textcolor[rgb]{0.7,0,0}{+ 2.30}) & 49.83 (\textcolor[rgb]{0.7,0,0}{-  6.83}) & 27.44 (\textcolor[rgb]{0.7,0,0}{- 2.31}) \\
       VDD-None & 7B &  62.97 (\textcolor[rgb]{0.7,0,0}{+ 6.01}) & \textbf{42.62} (\textcolor[rgb]{0.7,0,0}{+ 7.21}) & 66.50 (\textcolor[rgb]{0.7,0,0}{+ 2.50}) & 57.34 (\textcolor[rgb]{0,0.7,0}{+  0.86}) & 28.22 (\textcolor[rgb]{0.7,0,0}{- 1.53}) \\
      
      \rowcolor{blue!5} \textbf{w. \IFG} & 7B & \textbf{63.49} (\textcolor[rgb]{0.7,0,0}{+ 6.53}) & 39.40 (\textcolor[rgb]{0.7,0,0}{+ 3.99}) & \textbf{68.40} (\textcolor[rgb]{0.7,0,0}{+ 5.40}) & \textbf{29.14} (\textcolor[rgb]{0.7,0,0}{- 27.52}) & \textbf{15.62} (\textcolor[rgb]{0.7,0,0}{- 14.13}) \\
      \bottomrule  
    \end{tabular}  
  }  
  \caption{
  Comparison of \IFG with training-free methods designed to mitigate hallucinations under various decoding strategies. Performance gap compared to the base model(LLaVA-1.5) are noted in parentheses. \textcolor[rgb]{0.7,0,0}{Red} denotes improvements, ; \textcolor[rgb]{0,0.7,0}{green} indicates negative effects. Additional results for other model sizes are in \Cref{sec:sig_compare}.
  }  
  
  \label{tab:comparison}  
\end{table*}

\subsection{Evaluation Setup}

We conduct experiments across three categories:

\noindent\textbf{Visual Comprehension:} 
\textbf{MMBench}\citep{MMBench} evaluates fine-grained abilities of LVLMs, assessed with accuracy.
\textbf{TextVQA}~\citep{textvqa} employs VQA accuracy~\citep{VQA} as metric for questions with text within images. We send models with pure images for evaluation.
\textbf{MM-VET}~\citep{mmvet} evaluates LVLMs with GPT-4 in free-form question-answering.

\noindent\textbf{Open-ended Generation:} \textbf{LLaVA-Bench}~\citep{LLAVA} uses GPT-4 to compare generated answers with reference answers.

\noindent\textbf{Visual Hallucination:} 
\textbf{MMHal-Bench}~\citep{LLAVA-RLHF} evaluates hallucinations and response informativeness, with GPT-4 comparing model outputs to human responses and object labels.
\textbf{Object HallBench}~\citep{objhal} detects object hallucinations by comparing model outputs with COCO labels~\citep{COCO}. We follow same setup as \citep{RLHF-V}, which adds diverse prompts with detailed image descriptions for evaluations.

\subsection{Experimental Results}
We evaluate our method across benchmarks in \Cref{tab:comparison}, comparing with baseline models: (1) LVLMs after multimodal alignment training($\dag$); (2) training-free methods for mitigating hallucinations($\ddag$); and (3) reinforcement learning methods($\circ$).
\OURS consistently outperforms across all benchmarks. 
Notably, over LLaVA-1.5~\citep{LLAVA15}, which shares same training data and architecture, \OURS achieves double-digit percentage gains across different model sizes(indicated by $\Delta$), demonstrating strong scalability.
\OURS also surpasses training-free methods such as VCD~\citep{vcd}, VDD~\citep{alabdulmohsin2024fractalpatternsilluminatesuccess} and ICD~\citep{ICD}, achieving nearly 20 points reduction on Obj Hall. 
The underperformance of these methods further indicates that adding randomly generated noise on input images or simply remove images during the inference injects the unexpected information that was not present during training, thereby diminishing robustness of their methods.
Compared to reinforcement learning-based methods, which require extensive training resources and additional high-quality feedback data, \OURS remains effective and cost-efficient while delivering superior results. 
While RLHF-V achieves best score on Obj Hall, likely due to overfitting from overlap with its training data, base model, and benchmark~\citep{RLHF-V,MUFFIN}. 
In contrast, \OURS outperforms RLHF-V by a wide margin in other tasks (e.g., +32.9 on LLaVABench). 
Overall, our model demonstrates lower hallucination rates and higher visual comprehension scores without requiring additional resources, showcasing the effectiveness of our proposed method.
For thorough evaluations, we conduct experiments across various benchmarks in \Cref{sec:Additional_eval}, including ScienceQA~\citep{sciqa}, POPE~\citep{pope}, SeedBench~\citep{SEEDbench}, and MMVP~\citep{Eyes}, showing consistent improvements. We also perform \OURS on LLaVA-Next to demonstrate the generalization across different model architectures in \Cref{sec:llavanext}.

\subsection{Analysis Results}
\paragraph{Effect of \IFG in Decoding Perspective}
To distinguish \OURS from prior works, we investigate effectiveness of \IFG in different decoding strategies. As shown in \Cref{tab:comparison}, existing training-free methods, like VCD~\citep{vcd} and VDD-None~\citep{zhang2024debiasingmultimodallargelanguage}, only yield gains under \textbf{Nucleus Sampling}~\citep{Nucleus}, while \IFG consistently improves performance under both \textbf{Greedy} and \textbf{Nucleus Sampling}. 
It is further validated across different model sizes in \Cref{sec:sig_compare}.

VCD contrasts outputs from original and distorted visual inputs, while VDD uses text-only inputs. However, Adding random noise or omitting visual inputs at inference create discrepancies not present during training, leading to degraded performance and reduced robustness, especially on benchmarks like MMBench, where outputs are short and deterministic.
Greedy Sampling, which selects most probable token, offers limited tolerance for the introduced noise, making these methods less effective.
By contrast, Nucleus Sampling introduces randomness by sampling from a probability distribution, which mitigate sensitivity to noise, making these methods appear effective. However, this randomness may harm performance in tasks requiring precise outputs (e.g., multi-choice QA), often underperforming compared to Greedy Sampling.

In contrast, \IFG replaces visual inputs with a learnable soft visual prompt that preserves input patterns while compelling model to prioritize text inputs. It ensures consistency between training and inference, enabling \IFG to deliver robust gains under both decoding strategies.
Additional comparisons in \Cref{sec:sig_compare} further demonstrate \IFG's effectiveness against IBD~\citep{IBD}, ICD~\citep{ICD}, VDD-UNK~\citep{zhang2024debiasingmultimodallargelanguage}, and a variant using a blank image.


\begin{table}  
  \centering  
  \scriptsize  
  \resizebox{0.82\linewidth}{!}{  
    \begin{tabular}{@{}lcccc@{}}  
      \toprule  
      \multirow{2}{*}{Model} & \multicolumn{4}{c}{LLaVABench} \\
      \cmidrule(lr){2-5}  
       & Complex & Conv & Detail & All \\
      \midrule  
      LLaVA-1.5 & 75.50 & 54.10 & 56.60 & 64.40 \\
      w. FastV & 79.80 & 54.10 & 46.70 & 63.90 \\
     \rowcolor{blue!5}  $\Delta$  & \textcolor[rgb]{0.7,0,0}{+ 4.30} & \textcolor[rgb]{0,0.7,0}{+ 0.00} & \textcolor[rgb]{0,0.7,0}{- 9.90} & \textcolor[rgb]{0,0.7,0}{- 0.50} \\
      \midrule
      \MDA & 83.20 & 59.70 & 59.20 & 70.30 \\
      w. FastV & 10.70 & 10.20 & 10.40 & 10.50 \\
      \rowcolor{blue!5} $\Delta$ & \bf \textcolor[rgb]{0,0.7,0}{- 72.50} & \bf \textcolor[rgb]{0,0.7,0}{- 49.50} &\bf \textcolor[rgb]{0,0.7,0}{- 48.80} & \bf \textcolor[rgb]{0,0.7,0}{- 59.80} \\

      \bottomrule  
    \end{tabular}  
  }  
\caption{Performance on LLavaBench between LLaVA-1.5 and those with \MDA, with and without FastV. 
}
 
  \label{tab:MDA_comparison}  
\end{table}
\paragraph{How do LVLMs Treat Visual Inputs with \MDA?}
To evaluate the effectiveness of \MDA in mitigating language bias caused by training scale disparities, we analyze how LVLMs process visual inputs across layers. 
To assess whether \MDA addresses this issue, we adopt the pruning method proposed by~\citet{FastV} on LLaVA-1.5 with \MDA by pruning half of the visual tokens in deeper layers and measuring performance on LLaVA-Bench. Prior work~\citep{FastV} shows that pruning visual tokens in deeper layers has minimal impact on standard LVLMs, indicating poor utilization of visual inputs at those layers. In contrast, our results in \Cref{tab:MDA_comparison} show a significant performance drop when pruning is applied to the model with \MDA, confirming that visual information is effectively utilized throughout all layers—not just shallow ones.
\MDA ensures comprehensive attention to visual inputs across the model’s layers, thereby facilitating LVLMs in fully exploiting its visual comprehension capabilities. The 7.7-points improvement for complex tasks on LLaVABench in \Cref{tab:MDA_comparison} validate this conclusion, as complex tasks generally require deeper layers for precise understanding~\citep{benartzy2024attendfirstconsolidatelater,jin2024exploringconceptdepthlarge}.

\paragraph{Ablation Study}
To understand contributions of each component, we conduct an ablation study across multiple benchmarks in \Cref{tab:ablation} on the 7B model under different decoding strategies.
Removing \MDA (“ w/o \MDA”) causes a significant drop in performance, particularly on LLavaBench and MM-VET. This suggests that \MDA is crucial for enabling the model to effectively integrate visual information across the model. Excluding the \IFG(“ w/o \IFG”) also leads to a notable performance decrease across all benchmarks. 
Both components individually contribute to substantial improvements over the baseline LLaVA-1.5 model. Even when one component is removed, the model still outperforms the baseline. To further validate \OURS, we conduct ablation studies across various model sizes on multiple benchmarks in \Cref{sec:ablation_13b}.

\begin{table}  
  \centering  
  \small  
  \setlength{\tabcolsep}{4pt} 
  \resizebox{0.85\linewidth}{!}{ 
    \begin{tabular}{@{}llccc@{}}  
      \toprule  
      Sampling & Model & TextVQA & LLavaBench & MM-VET \\
      \midrule  
      \multirow{5}{*}{Greedy} & LLaVA-1.5 & 46.05 & 64.40 & 31.10 \\
                              & \cellcolor{blue!5} \bf \OURS  & \cellcolor{blue!5} \textbf{46.94} & \cellcolor{blue!5} \textbf{72.20} & \cellcolor{blue!5} \textbf{33.50} \\
                              & -w/o. \MDA & 46.77 & 70.60 & 32.00 \\
                              & -w/o. \IFG & 46.03 & 70.30 & 32.80 \\
      \midrule  
      \multirow{5}{*}{Nucleus} & LLaVA-1.5 & 35.41 & 63.00 & 29.80 \\
                               & \cellcolor{blue!5} \bf \OURS  & \cellcolor{blue!5} \textbf{42.05} & \cellcolor{blue!5} \textbf{72.20} & \cellcolor{blue!5} \textbf{35.20} \\
                               & -w/o. \MDA & 39.40 & 68.40 & 33.30 \\
                               & -w/o. \IFG & 36.40 & 67.80 & 30.50 \\
      \bottomrule  
    \end{tabular}  
  }  
  \caption{Ablation study on under different decoding strategy across multiple benchmarks on 7B model.}  
  \label{tab:ablation}  
\end{table}

\paragraph{Effectiveness on Different Model Architecture}
To validate robustness of \OURS, we conduct additional experiments on other model architectures. We use LLaVA-NEXT~\citep{liu2024llavanext} 
as base model, which supports dynamic resolution. Due to training data availability, we leverage training data from fully open-sourced version of LLaVA-NEXT~\citep{chen2024open}. 
Results show that our approach applies to LLaVA-NEXT as well, proving its versatility across different architectures and training methods. See \Cref{sec:llavanext} for details.

\paragraph{Effect of Bidirectional Attention in \MDA for Visual Inputs.}
To validate our design choice and highlight that the core strength of \MDA lies in its parallel dual-attention mechanism, we compare attention strategies for visual inputs in \Cref{sec:design_choice}. 
Results show that even with causal attention, \MDA outperforms the baseline, confirming the effectiveness of the dual-attention design. Bidirectional attention yields greater gains, aligning better with the spatial nature of visual data and justifying its use in \MDA.


\paragraph{Parameter-Efficient Tuning.}
While our primary focus is full-model retraining to ensure fair and rigorous comparisons across methods, we also explore a lightweight alternative through parameter-efficient tuning. Specifically, we apply our proposed method in \Cref{sec:ffn_training}, accompanied by a detailed discussion, to demonstrate its effectiveness.

\paragraph{Parameter Study.}
We conduct the parameter study in \Cref{sec:Hyperparameters} with the detailed discussion.

\paragraph{Human Evaluation and Case Study.}
The human evaluation on LLaVABench and a practical case study are detailed in \Cref{sec:Human} and \Cref{sec:cases}, respectively, demonstrating effectiveness of \OURS.

\section{Conclusion}

This paper tackles the language bias in LVLMs, which often leads to neglect of visual inputs and hallucinatory responses. We identify two primary sources of this bias: gap in training scales between the pretraining and multimodal alignment, and learned inference bias. To reduce language bias, we introduced Multimodal Dual-Attention Mechanism (\MDA) and Soft-Image Guidance (\IFG). \MDA enhances the integration of visual inputs across all layers. \IFG proposes a novel decoding strategy to mitigate over-reliance on adjacent text tokens, using a learnable soft visual prompt. 
Our work highlights the importance of addressing language biases from both training and inference perspectives, paving the way for more advanced LVLMs.


\section{Limitation}
\label{sec:limit}
 Despite the promising results demonstrated by \OURS in addressing the language bias of LVLMs,  several limitations must be acknowledged.
 First, although we validate the effectiveness of our method on two representative LVLMs that has different architecture—LLaVA-1.5 and LLaVA-Next—more extensive evaluation across a wider range of LVLM architectures is still lacking. This is primarily because our method targets the multimodal alignment stage that post-trains an LLM-based backbone into an LVLM, requiring fair comparisons that retrain models from scratch. However, for more advanced LVLMs such as Qwen-VL-2.5 and InternVL-3, the data and training details for their multimodal alignment stages are not fully open-sourced, making it infeasible to apply or evaluate our approach directly. 
 Nevertheless, language bias is commonly observed across various LVLMs~\citep{MMICL,MMSTAR,FastV} and even the SOTA LVLMs~\citep{wang2024qwen2vl} exhibits such phenomena. Therefore, inspired by this common observation and  the consistent gains observed across model sizes and different in our experiments, we anticipate the implementation and effectiveness of \OURS on diverse LVLMs. 
Additionally, due to resource constraints, we are unable to acquire LVLMs that achieve a similar scale of training between the LLM pretraining stages and the LVLM alignment stage to accurately validate the source of language bias. 
Finally, while \OURS has significantly reduced hallucinations in LVLMs and enhanced visual comprehension capabilities, there remains a possibility for it to produce hallucinations or disseminate misinformation. Therefore, it still should be employed with caution in critical applications.
Consequently, future research could involve broadening our approach to include a wider spectrum of LVLMs with different architectures and training them using a comparable training scale to observe the manifestations of language bias.

\section*{Acknowledgements}
We would like to thank the anonymous reviewers for their suggestions.
This work is supported by the National Science Foundation of China under Grant No.61876004.



\bibliography{main}

\clearpage
\newpage
\appendix
\section{Appendix}
\noindent This Appendix is organized as follows.  

\begin{itemize}[left=0pt]
    \item In \Cref{sec:Details}, we show implementation details of our method: training details(\Cref{sec:Training_Details}), datasets(\Cref{sec:datasets}) and hyperparameters(\Cref{sec:Hyperparameters}).
    \item In \Cref{app:related_work}, we present the related work of this paper, focusing on the language bias in LVLMs(\Cref{app:language_bias}) and the method to address such bias(\Cref{app:address_bias}).
    \item In \Cref{sec:Experimental_detail}, we present the details of our experiments and evaluation. Specifically, dataset and metric(\Cref{sec:Dataset_and_metric}), baselines(\Cref{sec:Baselines}) and GPT-4 Version(\Cref{sec:gpt4})
    \item In \Cref{sec:Additional}, we provide the additional experiments, including the evaluations across wide-range of benchmarks(\Cref{sec:Additional_eval}), baselines(\Cref{sec:sig_compare}), different architecture(\Cref{sec:llavanext}), different design choice(\Cref{sec:design_choice}), different model size(\Cref{sec:ablation_13b}) and parameter efficient training using \OURS~\Cref{sec:ffn_training}.
    \item In \Cref{sec:early-fusion}, we analyze early-fusion LVLMs like Chameleon, trained from scratch with a balanced mix of text and visual tokens, distinguishing them from the LVLMs discussed in this paper.
    \item In \Cref{sec:Parameter}, we detail the experiments and provide an in-depth discussion on the impact of hyperparameters, specifically the replace probability $\theta$(\Cref{sec:Replace_Probability}) and the scaling parameter $\lambda$(\Cref{sec:Scaling_Parameter}).
    
    \item In \Cref{sec:Human}, we present a human evaluation of \OURS versus LLaVA-1.5 across LLaVABench.  
    
    \item In \Cref{sec:cases}, we present more qualitative results.  
    
    \item In \Cref{sec:visualization}, we visualized the attention distribution across different layers in LLaVA-1.5 and \OURS.
\end{itemize}


\section{Training Details}
\label{sec:Details}
To make fair compression, we adopt the same training settings as LLaVA-1.5~\citep{LLAVA15}, maintaining consistency in hyperparameters, training dataset, data preprocessing, and model architecture. The only differences lies in the introduction of the multimodal dual-attention mechanism and the learnable soft visual prompt for soft-image guidance.

\subsection{Training}
\label{sec:Training_Details}

Following the setting of LLaVA-1.5~\citep{LLAVA15}, we employ CLIP-ViT-L-14-336~\citep{radford2021clip} as the visual encoder, paired with a two-layer MLP adapter to project visual embeddings from the encoder to the LLM backbone. Vicuna-1.5~\citep{vicuna2023} serves as the LLM backbone.
All of the experiments are conducted on the 8 $\times$ A100 GPUs, each with 40 GB of memory. We employ the Deepspeed Zero2~\citep{rajbhandari2020zero} and Deepspeed Zero3~\citep{rajbhandari2020zero} for training the 7B and 13B model, respectively.

In addition to these standard components of LLaVA-1.5, our method includes two significant modifications to the model architecture. Firstly, we adopt the multimodal dual-attention mechanism proposed in this paper, replacing the vanilla self-attention in the LLM. This modification slightly increases the computational cost due to the dual-attention calculation.
We further incorporate a learnable soft visual prompt for soft-image guidance.
We maintain a learnable embedding with dimensions
\([l_{\text{visual}}, h_{\text{LLM}}]\),
where \(l_{\text{visual}}\) is the visual embedding length and \(h_{\text{LLM}}\) is the LLM hidden state size.
In our practice, the learnable soft visual prompt has a size of $[576, 4096]$ for a 7B model and $[576, 5120]$ for a 13B model, which correspondingly adds 2.36M and 2.95M parameters to the 7B and 13B models. Compared to the billion-level parameters of these LVLMs, the additional parameters account for only 0.03\% and 0.02\%, respectively, which are minimal and negligible.
Therefore, compared to LLaVA-1.5, our method does not require additional training resources or computational costs, thereby demonstrating the efficiency of our approach. Practically speaking, the time cost of our method is approximately identical to that of LLaVA-1.5 under the same setting.
\begin{table}[t]  
\centering  
\resizebox{0.8\linewidth}{!}{%
\begin{tabular}{l|l}  
\toprule  
Dataset & Data Size \\ \midrule  
LLaVA~\citep{LLAVA} & 158K \\   
ShareGPT~\citep{sharegpt} & 40K \\   
VQAv2~\citep{goyal2017vqav2} & 83K \\   
GQA~\citep{hudson2019gqa} & 72K \\   
OKVQA~\citep{okvqa} & 9K \\   
OCRVQA~\citep{mishra2019ocrvqa} & 80K \\   
A-OKVQA~\citep{schwenk2022okvqa} & 66K \\   
TextCaps~\citep{sidorov2020textcaps} & 22K \\   
RefCOCO~\citep{kazemzadeh2014referitgame} & 48K \\   
VG~\citep{krishna2017visual} & 86K \\   
Total & 665K \\ \bottomrule  
\end{tabular}  
}  
\caption{  
Instruction-following Data Mixture Used for Finetuning~\citep{LLAVA15}.  
}  
\label{tab:data_mixture}  
\end{table}  

\subsection{Data}
\label{sec:datasets}

We strictly follows the data setting of LLaVA-1.5 for both pretraining and finetuning. Specifically, the LLaVA-558K~\citep{LLAVA} for pertraining and a mixture of instruction-following data for finetuning shown in \Cref{tab:data_mixture}.

\subsection{Hyperparameters}
\label{sec:Hyperparameters}

\begin{figure}[th]  
  \centering  
  \includegraphics[width=\linewidth]{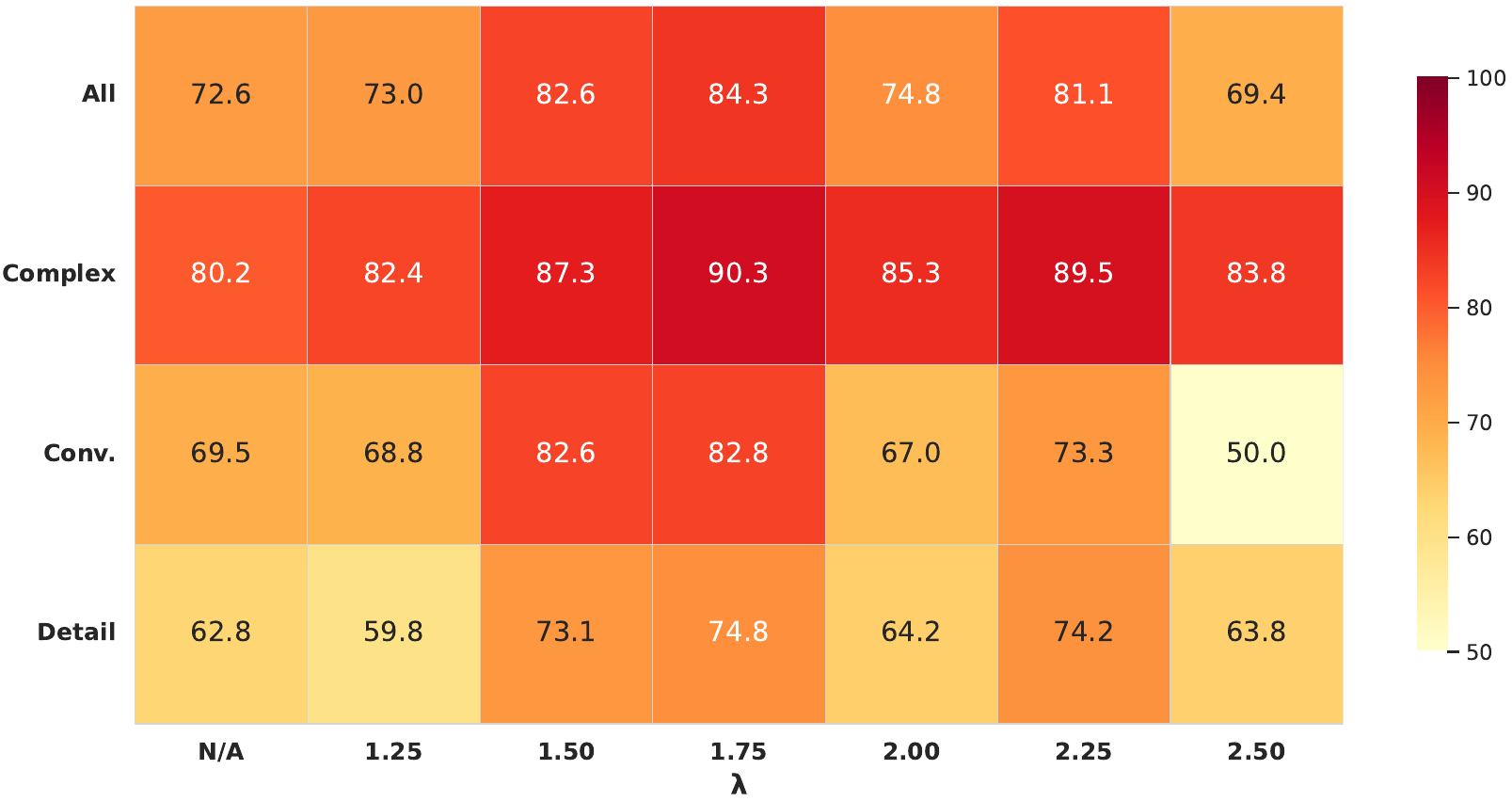}  
 \caption{Model performance on LLaVABench across various scaling parameter $\lambda$.}
  \label{fig:heat_llava}  
\end{figure}

We utilize the identical set of hyperparameters as the original LLaVA-1.5~\citep{LLAVA15}, with the exception of specifying the replacement probability $\theta$ for training with soft-image guidance. Detailed training hyperparameters for both stages are provided in \Cref{tab:hyperparameter}. 
During the inference, we use the hyperparameter $\lambda$ to control the guidance of the visual inputs on the response generation.  As illustrated in \Cref{fig:heat_llava}, we report the performance of the 13B model on LLaVABench across various the scaling parameter $\lambda$, thereby demonstrating the impact of different $\lambda$ scales on model performance. The optimal performance of our method under various $\lambda$ values is reported in the experiments.

\begin{table}[t]  
\centering  
\scalebox{0.95}{  
\begin{tabular}{l|c c}  
\toprule  
Hyperparameter & Pretrain & Finetune \\
\midrule  
batch size & 256 & 128 \\
lr & 1e-3 & 2e-5 \\
lr schedule & cosine decay & cosine decay \\
lr warmup ratio & 0.03 & 0.03 \\
weight decay & 0& 0 \\
optimizer & AdamW & AdamW \\
DeepSpeed stage & 2 & 3 \\
replace prob. $\theta$ & 10\% & 10\% \\
\bottomrule  
\end{tabular}  
}  
\caption{  
\textbf{Hyperparameters} of ~\OURS, which are the same as the original LLaVA-1.5~\citep{LLAVA15}, except that we set the replace probability $\theta$ for training with soft-image guidance.  
}  
\label{tab:hyperparameter}  
\end{table}

\section{Related Work}
\label{app:related_work}
\subsection{Language Bias in LVLMs}
\label{app:language_bias}
Despite the impressive capabilities of LVLMs~\citep{gpt4o,gemini1,MM1,wang2024qwen2vl,Llava-onevision,wu2025longvitu,chen2025multimodalrepresentationalignmentimage,zhao2025mentorefficientmultimodalconditionedtuning}, similar like the hallucination~\citep{llmHall,si2025teachinglargelanguagemodels} of LLMs~\citep{dubey2024llama3,gpt4o,gemini1,zhao2024mitigatinglanguagelevelperformancedisparity,an2025rethinkingsemanticparsinglarge}, these models still struggle with generating responses irrelevant to the context images~\citep{lan2024surveyhallucinationlargevisual,liu2024surveyhallucinationlargevisionlanguage}, e.g., hallucinating non-existent objects~\citep{zhou2024analyzingmitigatingobjecthallucination}.
\citet{MMICL} first identify this issue in LVLMs and name it as \textit{language bias}, i.e., LVLMs often ignore visual inputs and solely rely on text inputs, leading to hallucinations.
Similarly, \citet{MMSTAR} observe that LVLMs can derive answers directly from the world knowledge embedded in LLMs or deduce them solely from the textual input, even in the absence of visual information. 
\citet{FastV} further analyzes the attention distribution of prominent LVLMs, revealing an inefficient attention mechanism wherein attention computation over visual inputs is extremly inefficient in the deeper layers of LVLMs. 
Moreover, \citet{zhang2024debiasingmultimodallargelanguage} note that LVLMs tend to allocate more attention to text inputs and increasingly prioritize them, with attention to visual inputs diminishing as the length of generated text increases. 
%
These findings collectively indicate that LVLMs assign disproportionately low attention to visual inputs, limiting their ability to effectively utilize image information.




\subsection{Addressing Language Bias in LVLMs} 
\label{app:address_bias}

Given the language bias of LVLMs, they exhibit similar hallucination issues as LLMs~\citep{si2025gateauselectinginfluentialsamples,llmHall,si2025aligninglargelanguagemodels}, as well as modality-specific hallucinations such as object hallucination~\citep{objhal,pope}. 
As noted by~\citet{vcd}, this stems from the dominant influence of the LLM's pretraining distribution, making hallucination a prominent symptom of language bias.
Recent efforts have been proposed to mitigate the hallucination in LVLMs. LRV~\citep{lrv} attempts to apply supervised fine-tuning on a well-designed visual preference dataset. LLaVA-BPO~\citep{llavaBPO} proposes pipeline to gather preference datasets for preference learning to mitigate hallucination. Additionally, LLaVA-RLHF~\citep{LLAVA-RLHF} and RLHF-V~\citep{RLHF-V} introduce reinforcement learning for LVLM to reduce hallucinations. 
While effective, these methods typically necessitate substantial training data and computational resources. To address this, training-free methods have been proposed, including VCD~\citep{vcd}, IBD~\citep{IBD}, VDD~\citep{zhang2024debiasingmultimodallargelanguage}, and ICD~\citep{ICD}. These methods contrast outputs with those from image-free inputs (or with distorted images) to reduce influence of textual LLMs. However, these methods may introduce inconsistencies between training and inference, limiting their effectiveness.

\section{Detailed Experimental Settings}
\label{sec:Experimental_detail}

\begin{table*}[ht]
    \centering
    \scriptsize

    \resizebox{0.85\linewidth}{!}{%
    \begin{tabular}{@{}lcccccccc@{}}  
        \toprule  
        \multirow{1}{*}{Model} & \multirow{1}{*}{SCIQA$\uparrow$} & \multicolumn{3}{c}{POPE$\uparrow$} & \multicolumn{3}{c}{SeedBench$\uparrow$} & \multirow{1}{*}{MMVP$\uparrow$} \\
        \midrule  
        LLaVA-1.5 & 70.12 & 87.38 & 84.26 & 86.21 & 58.60 & 66.10 & 37.30 & 26.00 \\
        VCD          & 70.12 & 87.39 & 84.25 & 86.21 & 59.93 & 65.62 & 38.41 & 26.00 \\
       \rowcolor{blue!5} \OURS         & \textbf{71.26} & \textbf{87.74} & \textbf{85.60} & \textbf{86.50} & \textbf{61.35} & \textbf{67.46} & \textbf{38.19} & \textbf{32.00} \\
        \bottomrule  
    \end{tabular}
    }
    \caption{Experiments with more benchmarks across 7B model}
    \label{tab:add_comparison}
\end{table*}

\subsection{Dataset and Metric}
\label{sec:Dataset_and_metric}


\paragraph{MMBench}~\citep{MMBench} provides a progressive evaluation framework, advancing from perception to reasoning, and covers 20 fine-grained abilities. It is assessed through multiple-choice question answering, using accuracy as the metric.

\paragraph{MMBench}~\citep{MMBench} provides a progressive evaluation framework, advancing from perception to reasoning, and covers 20 fine-grained abilities. It is assessed through multiple-choice question answering, using accuracy as the metric.

\paragraph{TextVQA}~\citep{textvqa} is designed for visual question answering involving text within images. It employs VQA accuracy as the evaluation metric. Unlike LLaVA-1.5~\citep{LLAVA15}, which includes OCR results of the images in the question, our approach provides the model solely with the image and the question. This setup aims to assess the model's visual comprehension abilities without supplementary textual data.

\paragraph{MM-VET}~\citep{mmvet} evaluates multimodal understanding across six core vision-language capabilities over 128 tasks. The evaluation is conducted using GPT-4 to assess model performance in a free-form question-answering format. MM-Vet defines 16 integrations derived from combinations of these core capabilities, providing a structured assessment of models' abilities to handle complex multimodal tasks.

\paragraph{LLaVABench}~\citep{LLAVA} is utilized for evaluating open-ended generation capabilities. This benchmark consists of 60 tasks focused on LLaVA’s visual instruction-following and question-answering abilities in natural environments. It employs GPT-4 as the evaluator to compare the model's generated answers with reference answers, ensuring a comprehensive assessment of the model's generative performance.

\paragraph{Object HalBench} ~\citep{objhal} detects object hallucinations by comparing model outputs with COCO image labels~\citep{COCO}. \citet{RLHF-V} further augment this benchmark by adding eight diverse prompts with detailed image descriptions for stable evaluations. We follow the same evaluation setup and use GPT-4 as the evaluator. We report the two metrics in this benchmark: The response-level hallucination rate and the object-level hallucination rate.

\paragraph{MMHall-Bench}~\citep{LLAVA-RLHF} evaluates hallucinations and response informativeness. It employs GPT-4 to compare model output with human response and several object labels to get the final scores.

\subsection{Baselines}
\label{sec:Baselines}

\paragraph{General LVLMs} that have undergone multimodal alignment training. Specifically, we utilize LLaVA~\citep{LLAVA}, Qwen VL~\citep{Qwen-VL}, LLaVA-1.5~\citep{LLAVA15}, Muffin~\citep{MUFFIN}, and LRV~\citep{lrv} as representative baselines. These LVLMs are predominantly trained with multimodal data for alignment~\citep{LLAVA,Qwen-VL,MUFFIN} and fine-tuned using high-quality instruction data~\citep{LLAVA15,lrv}, thereby achieving exceptional performance in various multimodal tasks. For example, LRV~\citep{lrv} employs supervised fine-tuning on an expertly crafted visual preference dataset to mitigate hallucinations in LVLMs. Typically, these models integrate a pre-trained visual encoder with a large language model through an alignment module.

\begin{table}
  \centering  
  \scriptsize  
  \resizebox{0.7\columnwidth}{!}{  
    \begin{tabular}{lcc}  
      \toprule  
      Model & MMBench & TextVQA \\
      \midrule  
      \multicolumn{3}{c}{\cellcolor{olive!10}\textbf{Greedy Sampling}} \\
      \midrule  
      LLaVA-1.5~\citep{LLAVA15} & 64.61 & 46.05 \\
      -w. Two epoch & 65.63 & 45.83 \\
      \midrule  
      w. \IFG & 66.92 & 46.77 \\
      -w. Two epoch & 66.58 & 47.15 \\
      \midrule  
      \multicolumn{3}{c}{\cellcolor{olive!10}\textbf{Nucleus Sampling}} \\
      \midrule  
      LLaVA-1.5~\citep{LLAVA15} & 56.96 & 35.41 \\
      -w. Two epoch & 60.82 & 36.70 \\
      \midrule  
      w. \IFG & 63.49 & 39.40 \\
      -w. Two epoch & 62.97 & 41.27 \\
      \bottomrule  
    \end{tabular}  
  }  
  \caption{Performance comparison of models undergoes training for one or two epochs across MMBench and TextVQA.}  
  \label{tab:performance_comparison}  
\end{table}

\paragraph{Training-free methods} designed to mitigate hallucination of LVLMs. 
VCD~\citep{vcd} contrast model outputs generated from original inputs and distorted visual input to reduce over-reliance on statistical bias and unimodal priors. 
Similiarly, VDD~\citep{zhang2024debiasingmultimodallargelanguage}  contrast model outputs  from original inputs and inputs without visual inputs to  reduce the influence of textual LLMs. 
 OPERA~\citep{OPERA} introduces a penalty term on the model logits during the beam-search decoding to mitigate the over-trust toward a few summary tokens.
 Less-is-more~\citep{Less} proposes a selective end-of-sentence (EOS) special token supervision loss coupled with a data filtering strategy to improve the model’s capacity for timely termination of generation, thereby mitigating hallucinations.

\paragraph{Reinforcement Learning-based method} aimed at aligning LVLM outputs with human intentions to mitigate hallucination of LVLMs. 
Specifically, POVID~\citep{POVID} addresses hallucinations in VLLMs using AI-generated feedback. It first prompts GPT-4V to add  hallucinations to correct answers and use distorts images to  invoke the VLLM's inherent hallucination tendencies. The model is then trained with this generated data using direct preference optimization approaches~\citep{dpo} to mitigate hallucinations.
HA-DPO~\citep{HA-DPO} propose a pipeline for constructing positive and negative sample pairs and adopt the direct preference optimization~\citep{dpo} using the constructed dataset to reduces hallucination.
RLHF-V~\citep{RLHF-V} employs the Muffin~\citep{MUFFIN} as the LLM backbone and collects 1.4k fine-grained correctional human feedback.
The model is trained using this dataset through the proposed dense direct preference optimization method to reduce hallucination.
LLaVA-BPO~\citep{llavaBPO} proposes a pipeline to gather preference datasets and conduct preference learning to mitigate this type of hallucination.

\begin{table*}[htbp]
    \centering
    \scriptsize
    \resizebox{0.7\linewidth}{!}{%
        \begin{tabular}{@{}lcccccc@{}}
            \toprule
            \multirow{2}{*}{Method} & \multirow{2}{*}{LLaVABench$\uparrow$} & \multirow{2}{*}{MM-VET$\uparrow$} & \multicolumn{2}{c}{MMHall} & \multicolumn{2}{c}{Obj Hall} \\
            \cmidrule(lr){4-5} \cmidrule(lr){6-7}  
            &  &  & Score $\uparrow$ & Hall $\downarrow$ & Res $\downarrow$ & Obj $\downarrow$ \\
            \midrule
            LLaVA-1.5~\citep{LLAVA15} & 64.40 & 31.10 & 2.19 & 59 & 46.71 & 25.08 \\
            IBD~\citep{IBD}          & 64.60 & 31.10 & 2.24 & 58 & 46.31 & 24.16 \\
            ICD~\citep{ICD}          & 64.70 & 31.10 & 2.18 & 59 & 47.40 & 25.00 \\
            VDD-UNK~\citep{zhang2024debiasingmultimodallargelanguage}      & 65.30 & 31.00 & 2.22 & 56 & 46.71 & 24.82 \\
            \IFG-blank    & 68.40 & 31.50 & 2.42 & 52 & 34.41 & 17.80 \\
            \rowcolor{blue!5} \bf \IFG      & \textbf{70.60} & \textbf{32.00} & \textbf{2.47} & \textbf{50} & \textbf{30.36} & \textbf{15.16} \\
            \bottomrule
        \end{tabular}
    }
    \caption{Comparison of SIG with other baselines on 7B model}
    \label{tab:comparison_sig_more}
    
\end{table*}

\subsection{GPT-4 Version}
\label{sec:gpt4}

For all evaluations conducted using the GPT-4(evaluation for Object HalBench, MMHall-Bench, LLaVABench, and MM-VET), we utilized the \texttt{GPT-4 API} in October 2024. It ensures consistency with prior research \citep{mmvet,RLHF-V,LLAVA-RLHF,LLAVA}. According to the documentation provided by OpenAI\footnote{https://platform.openai.com/docs/models/gpt-4-turbo-and-gpt-4}, \texttt{GPT-4 API} currently points to \texttt{GPT-4-0613 API}.

\begin{figure}  
  \centering  
  \includegraphics[width=\linewidth]{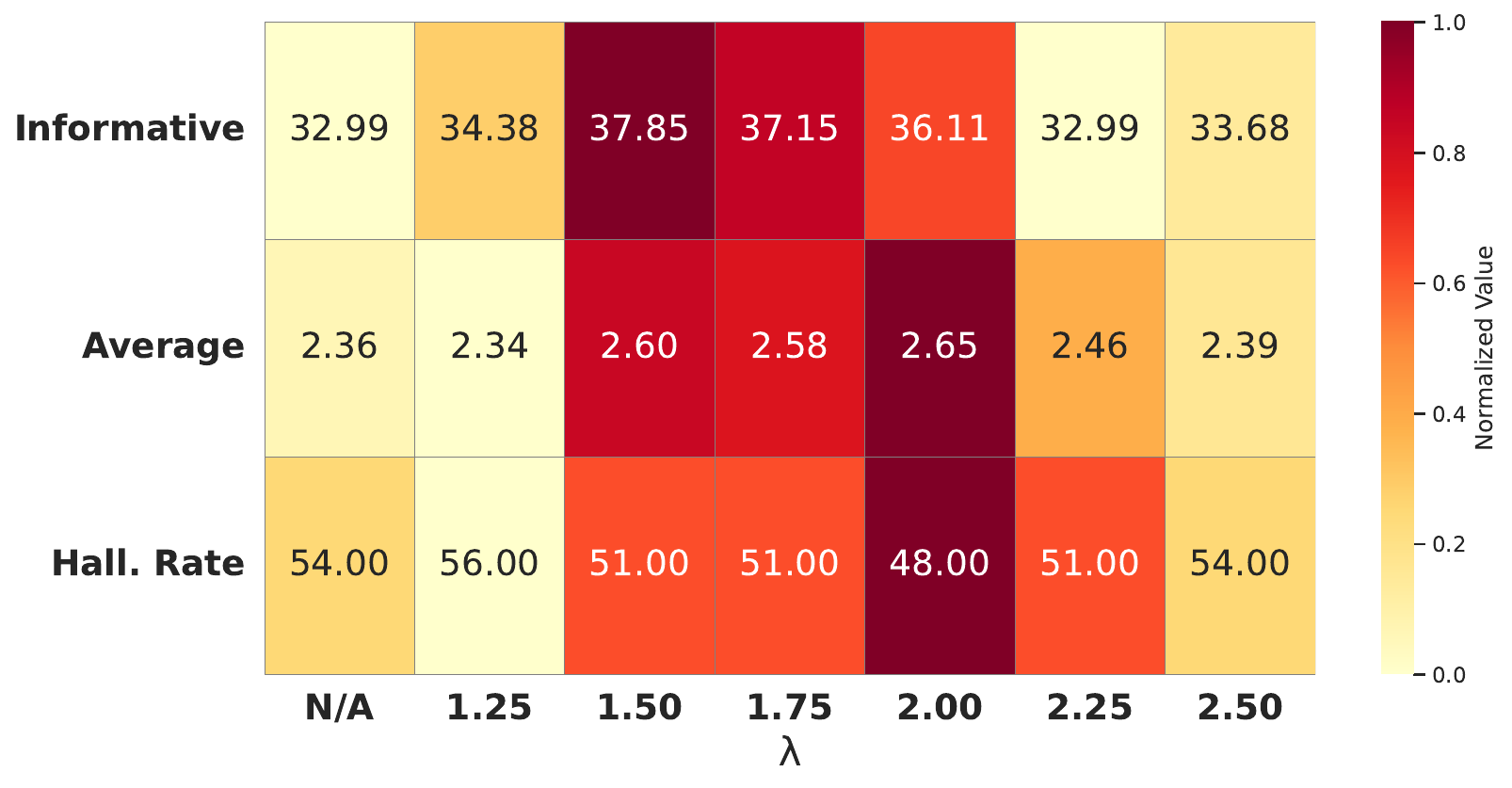}  
 \caption{Model performance on MMHall-Bench across various scaling parameter $\lambda$.}
  \label{fig:heat_hall}  
\end{figure}

\section{Additional Experiments}
\label{sec:Additional}

\subsection{Additional Evaluations across other benchmarks}
\label{sec:Additional_eval}

To further demonstrate the generalizability of \OURS, we conducted experiments on additional benchmarks, including ScienceQA, POPE, SeedBench, and MMVP. The results presented in \Cref{tab:add_comparison} consistently show improvements, confirming the effectiveness of our method.

\subsection{Comparison Between \IFG and Other Methods}
\label{sec:sig_compare}

\Cref{tab:comparison_sig_more} compares \IFG with other training-free baselines, including a variant using a blank image instead of the learnable soft-image prompt. The results show that \IFG outperforms all baselines, with the learnable prompt significantly surpassing the blank-image variant while adding only 0.02--0.03\% more parameters.

\Cref{tab:comparison_13B} compares \IFG with other training-free baselines for the 13B model. The results confirm that while prior training-free approaches improve performance only with Nucleus Sampling, \IFG demonstrates effectiveness across all decoding settings.

\begin{table*} [t]
  \centering  
  \scriptsize  

  \resizebox{0.8\textwidth}{!}{  
    \begin{tabular}{@{}lcccccc@{}}  
      \toprule  
      \multirow{2}{*}{Method} & \multirow{2}{*}{Model Size} & \multirow{2}{*}{MMBench$\uparrow$} & \multirow{2}{*}{TextVQA$\uparrow$} & \multirow{2}{*}{LLaVABench$\uparrow$} & \multicolumn{2}{c}{Obj Hall} \\   
      \cmidrule(lr){6-7}  
      & & & & & Res $\downarrow$ & Obj $\downarrow$ \\   
      \midrule  
      \multicolumn{7}{c}{\cellcolor{olive!10}\textbf{Greedy Sampling}} \\   
      \midrule  
      LLaVA-1.5 & 13B & 67.74 & 48.66 & 72.50 & 47.06 & 23.33 \\
      VCD & 13B & 68.38 (\textcolor[rgb]{0.7,0,0}{+ 0.64}) & 48.63 (\textcolor[rgb]{0,0.7,0}{- 0.03}) & 73.60 (\textcolor[rgb]{0.7,0,0}{+ 1.10}) & 46.37 (\textcolor[rgb]{0.7,0,0}{- 0.69}) & 23.10 (\textcolor[rgb]{0.7,0,0}{- 0.23}) \\
       VDD-None & 13B & 68.56 (\textcolor[rgb]{0.7,0,0}{+ 0.82}) & 47.31 (\textcolor[rgb]{0,0.7,0}{- 1.35}) & 73.00 (\textcolor[rgb]{0.7,0,0}{+ 0.05}) & 44.64 (\textcolor[rgb]{0.7,0,0}{- 2.42}) & 22.23 (\textcolor[rgb]{0.7,0,0}{- 1.10}) \\
      
      \rowcolor{blue!5}\textbf{w. \IFG} & 13B & \textbf{70.19} (\textcolor[rgb]{0.7,0,0}{+ 2.45}) & \textbf{48.74} (\textcolor[rgb]{0.7,0,0}{+ 0.07}) & \textbf{74.70} (\textcolor[rgb]{0.7,0,0}{+ 2.20}) & \textbf{28.27} (\textcolor[rgb]{0.7,0,0}{- 18.79}) & \textbf{15.21} (\textcolor[rgb]{0.7,0,0}{- 8.12}) \\
      \midrule  
      \multicolumn{7}{c}{\cellcolor{olive!10}\textbf{Nucleus Sampling}} \\   
      \midrule  
      LLaVA-1.5 & 13B & 62.11 & 38.92 & 68.10 & 50.52 & 25.74 \\
      VCD & 13B & 65.38 (\textcolor[rgb]{0.7,0,0}{+ 3.27}) & 43.56 (\textcolor[rgb]{0.7,0,0}{+ 4.64}) & 70.70 (\textcolor[rgb]{0.7,0,0}{+ 2.60}) & 49.83 (\textcolor[rgb]{0.7,0,0}{- 0.69}) & 24.23 (\textcolor[rgb]{0.7,0,0}{- 1.51}) \\
      
       VDD-None & 13B & \textbf{66.32} (\textcolor[rgb]{0.7,0,0}{+ 4.21}) & \textbf{45.99} (\textcolor[rgb]{0.7,0,0}{+ 7.07}) & 71.40 (\textcolor[rgb]{0.7,0,0}{+ 3.30}) & 47.90 (\textcolor[rgb]{0.7,0,0}{- 2.62}) & 23.25 (\textcolor[rgb]{0.7,0,0}{- 2.49}) \\
            
      \rowcolor{blue!5}\textbf{w. \IFG} & 13B & 64.77 (\textcolor[rgb]{0.7,0,0}{+ 2.66}) & 40.31 (\textcolor[rgb]{0.7,0,0}{+ 1.39}) & \textbf{72.00} (\textcolor[rgb]{0.7,0,0}{+ 3.90}) & \textbf{30.55} (\textcolor[rgb]{0.7,0,0}{- 19.97}) & \textbf{17.45} (\textcolor[rgb]{0.7,0,0}{- 8.29}) \\
      \bottomrule  
    \end{tabular}  
  }  
  \caption{
  Comparison of \IFG with training-free methods under different decoding strategies in 13B model. Performance gap compared to the base model(LLaVA-1.5) are noted in parentheses. \textcolor[rgb]{0.7,0,0}{Red} denotes positive improvements, while \textcolor[rgb]{0,0.7,0}{green} indicates negative effects.}  
  
  \label{tab:comparison_13B}  
\end{table*}

\subsection{Evaluation Across Different Model Architectures}
\label{sec:llavanext}
To ensure a fair comparison, we train the LVLM from scratch using our method and evaluate its performance against baseline models. Given the availability of training data, we select LLaVA-1.5~\citep{LLAVA15} as our base model and strictly adhere to its training settings, including the same dataset and model backbone.
To further validate the robustness of our approach, we conduct additional experiments across various model architectures. Specifically, we use LLaVA-NEXT~\citep{liu2024llavanext} as the base model, which supports dynamic resolution. Due to training data availability, we leverage the dataset from the fully open-sourced version of LLaVA-NEXT~\citep{chen2024open} and adhere to its training settings. We set the Vicuna-1.5~\citep{vicuna2023} language model backbone and ViT-L-14-336~\citep{radford2021clip} as the visual encoder.
Our preliminary results, presented in \Cref{tab:llava_next}, indicate that similar performance trends hold across additional LVLMs. This underscores that our approach is not limited to a specific architecture or training setup.



\begin{table}[ht]  
  \centering  
  \scriptsize  
  \resizebox{0.95\linewidth}{!}{  
    \begin{tabular}{@{}lcccccc@{}}  
        \toprule  
      \multirow{2}{*}{Model} & \multicolumn{2}{c}{Obj Hall} & \multicolumn{3}{c}{MMHall} & MM-VET $\uparrow$ \\
      \cmidrule(lr){2-3} \cmidrule(lr){4-6}  
       & Res $\downarrow$ & Obj $\downarrow$ & Score $\uparrow$ & Hall $\downarrow$ & & \\
      \midrule  
      LLaVA-Next & 13.81 & 7.50 & 2.67 & 51.00 & & 37.6 \\
      \rowcolor{blue!5} \bf \OURS & \textbf{7.92} & \textbf{4.29} & \textbf{2.84} & \textbf{49.00} & & \textbf{42.2} \\
      \bottomrule  
    \end{tabular}  
  }  
  \caption{Performance of \OURS on LLaVA-Next.}
  \label{tab:llava_next}  
\end{table}

\subsection{Comparison of Different Attention Mechanism for Visual Inputs in \MDA}
\label{sec:design_choice}

\begin{table}[ht]
  \centering
  \small
  \begin{tabular}{@{}lcc@{}}
    \toprule
    \textbf{Method} & \textbf{MM-VET} $\uparrow$ & \textbf{LLavaBench} $\uparrow$ \\
    \midrule
    LLaVA-1.5 & 31.10 & 64.40 \\
    Causal Attn. & 31.90 & 69.60 \\
   \rowcolor{blue!5}  Bi-Attn.(\MDA) & \textbf{32.80} & \textbf{70.30} \\
    \bottomrule
  \end{tabular}
  \caption{Comparison of different visual attention strategies in \MDA.}
  \label{tab:mda_design}
\end{table}

To validate our design choice and highlight that the core strength of \MDA lies in its parallel dual-attention mechanism, we compare different attention strategies for visual inputs in \Cref{tab:mda_design}. The results show that even when using only causal attention, \MDA still yields performance gains over the baseline, confirming the effectiveness of the dual-attention design. However, bidirectional attention achieves more significant improvements, aligning more naturally with the spatial characteristics of visual data. This further supports our motivation for adopting bidirectional attention for visual inputs in \MDA.

\subsection{Ablation Studies Across Different Model Size}
\label{sec:ablation_13b}
To further validate our method, we conduct ablation studies across various model sizes on multiple benchmarks. Specifically, we perform an ablation study on the 13B model across multiple benchmarks to analyze the impact of different components.
\Cref{tab:13B_abliation} presents the results, demonstrating that our approach outperforms the baseline and its ablated variants across both MMBench and LLaVABench, under both greedy decoding and sampling strategies.

\begin{table}[htbp]
    \centering
    \resizebox{\linewidth}{!}{%
    \begin{tabular}{lcccc}
        \toprule
        & \multicolumn{2}{c}{MMBench} & \multicolumn{2}{c}{LLaVABench} \\
        \cmidrule(lr){2-3} \cmidrule(lr){4-5}
        & Greedy & Sampling & Greedy & Sampling \\
        \midrule
        LLaVA-1.5 & 67.74 & 62.11 & 72.5 & 68.1 \\
        w.o. \IFG      & 68.73 & 65.55 & 76.7 & 75.5 \\
        w.o. \MDA      & 68.99 & 64.77 & 74.7 & 72.0 \\
     \rowcolor{blue!5}   \OURS          & \textbf{70.01} & \textbf{66.92} & \textbf{78.5} & \textbf{76.6} \\
        \bottomrule
    \end{tabular}%
    }
    \caption{Ablation study on 13B models.}
    \label{tab:13B_abliation}
    
\end{table}

\subsection{Parameter-Efficient Tuning with \OURS}
\label{sec:ffn_training}

While our primary focus has been on full-model retraining to ensure fair and rigorous comparisons across methods, we also recognize the importance of lightweight and practical alternatives. Inspired by prior works~\citep{rome} highlighting the role of Feed-Forward Networks (FFN) in retaining knowledge for large language models, we explore a targeted fine-tuning strategy where only the FFN layers are updated. This design reduces the number of trainable parameters to approximately 7\% of the full model (500M vs.\ 7B), thus offering a computationally efficient alternative.

We conduct experiments on LLaVA-1.5~\citep{LLAVA15}, comparing the baseline, our full retraining method (LACING), and the FFN-only fine-tuning variant. As shown in \Cref{tab:ffn_finetune}, this lightweight strategy achieves performance that is highly competitive with full retraining, while requiring significantly fewer computational resources. Specifically, FFN-only fine-tuning yields only marginally lower performance than full retraining, underscoring the robustness of our approach even under practical constraints.

\begin{table}[ht]
  \centering
  \small
\resizebox{\linewidth}{!}{%
  \begin{tabular}{@{}lcc@{}}
    \toprule
    \textbf{Method} & \textbf{LLaVABench} $\uparrow$ & \textbf{MM-VET} $\uparrow$ \\
    \midrule
    LLaVA-1.5 & 64.40 & 31.10 \\
    \rowcolor{blue!5} \OURS (Full Retrain) & \textbf{72.20} & \textbf{35.20} \\
    \OURS (FFN-tuning) & 71.20 & 34.00 \\
    \bottomrule
  \end{tabular}
  }
  \caption{Performance comparison of full retraining and FFN-only-tuning strategies using \OURS on LLaVA-1.5.}
  \label{tab:ffn_finetune}
\end{table}

These findings demonstrate that LACING remains highly effective even when applied in a lightweight fine-tuning setting, providing strong empirical evidence for its practicality in resource-constrained scenarios. We will further elaborate on these results in the main revision.

\section{Analysis of Early-fusion LVLMs}
\label{sec:early-fusion}
\begin{figure}[t]  
  \centering  
    \includegraphics[width=\linewidth]{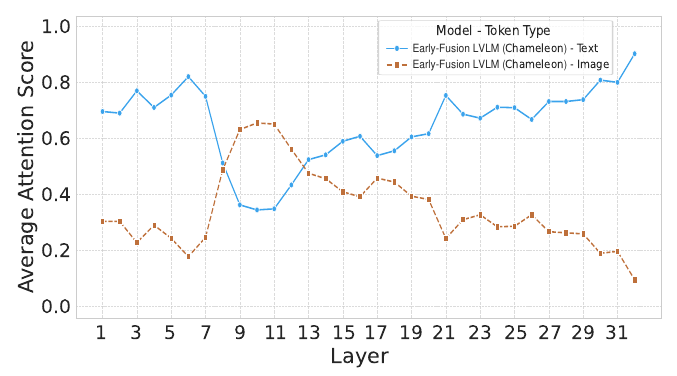}
  \caption{Average attention scores for output tokens towards text and visual tokens across different layers of early-fusion LVLMs (Chameleon~\citep{Chameleon}).}
  
  \label{fig:avg_attention_Chameleon}  
\end{figure}

The performance of LVLMs is often hindered by the disparity in training scales between the LLM pretraining phase and the subsequent LVLM alignment stage. This imbalance results in suboptimal utilization of visual inputs, as evidenced by the attention distributions: only the initial layers demonstrate significant attention to visual tokens, while the deeper layers tend to neglect them. In contrast, early-fusion LVLMs, such as Chameleon~\citep{Chameleon}, which are trained from scratch using a balanced mix of visual and textual tokens, exhibit a more effective modality fusion. As shown in \Cref{fig:avg_attention_Chameleon}, this balanced training approach enables the model to allocate attention more uniformly across modalities, thereby mitigating the issues associated with scale disparities during pretraining and alignment.

Following pervious work~\citep{MMICL}, we measure performance gaps on image-required vs.non-image-required questions gathered from Science QA~\citep{sciqa} dataset to evaluate language bias. As shown in \Cref{tab:visual_gap}, although showing better fusion, Chameleon, as well as other LVLMs still remains substantial language bias.

\section{Parameter Study}
\label{sec:Parameter}

\begin{table}[htbp]
    \centering
    
    \resizebox{0.7\linewidth}{!}{%
    \begin{tabular}{lccc}
        \toprule
        Model & Don’t Req & Req & Gap \\
        \midrule
        LLaVA       & 56.78 & 72.84 & 16.06 \\
        EVE         & 68.13 & 45.33 & 22.80 \\
        Chameleon   & 56.12 & 37.33 & 18.79 \\
        \bottomrule
    \end{tabular}%
    }
    \caption{Language Bias Evaluation.}
    \label{tab:visual_gap}
\end{table}

\subsection{Influence of the Replace Probability $\theta$}
\label{sec:Replace_Probability}
In the soft-image guidance we proposed, we intermittently replace the visual input with a learnable soft visual prompt at a predetermined probability rate to give the model an input without visual input during training. This introduces segments of training data that remain unseen by the LVLMs during training. Consequently, we make the model that undergoes training for two epochs as a baseline to ensure comprehensive exposure to all samples in the training dataset. Subsequently, we evaluate the model after one and two epochs of training on the same benchmarks to determine the impact of visual input replacement. 
The results presented in \Cref{tab:performance_comparison} indicate that neither the number of training epochs nor the visual input replacement significantly impacts model performance, as it remains consistent across various settings and does not exhibit a clear trend of performance variation related to different training settings.
To further establish the appropriate value of the  replace probability $\theta$, we present an experiment in \Cref{tab:theta} to identify the optimal value for this parameter.

\subsection{Impact of the Scaling Parameter $\lambda$}
\label{sec:Scaling_Parameter}

\begin{figure}
  \centering  
  \includegraphics[width=\linewidth]{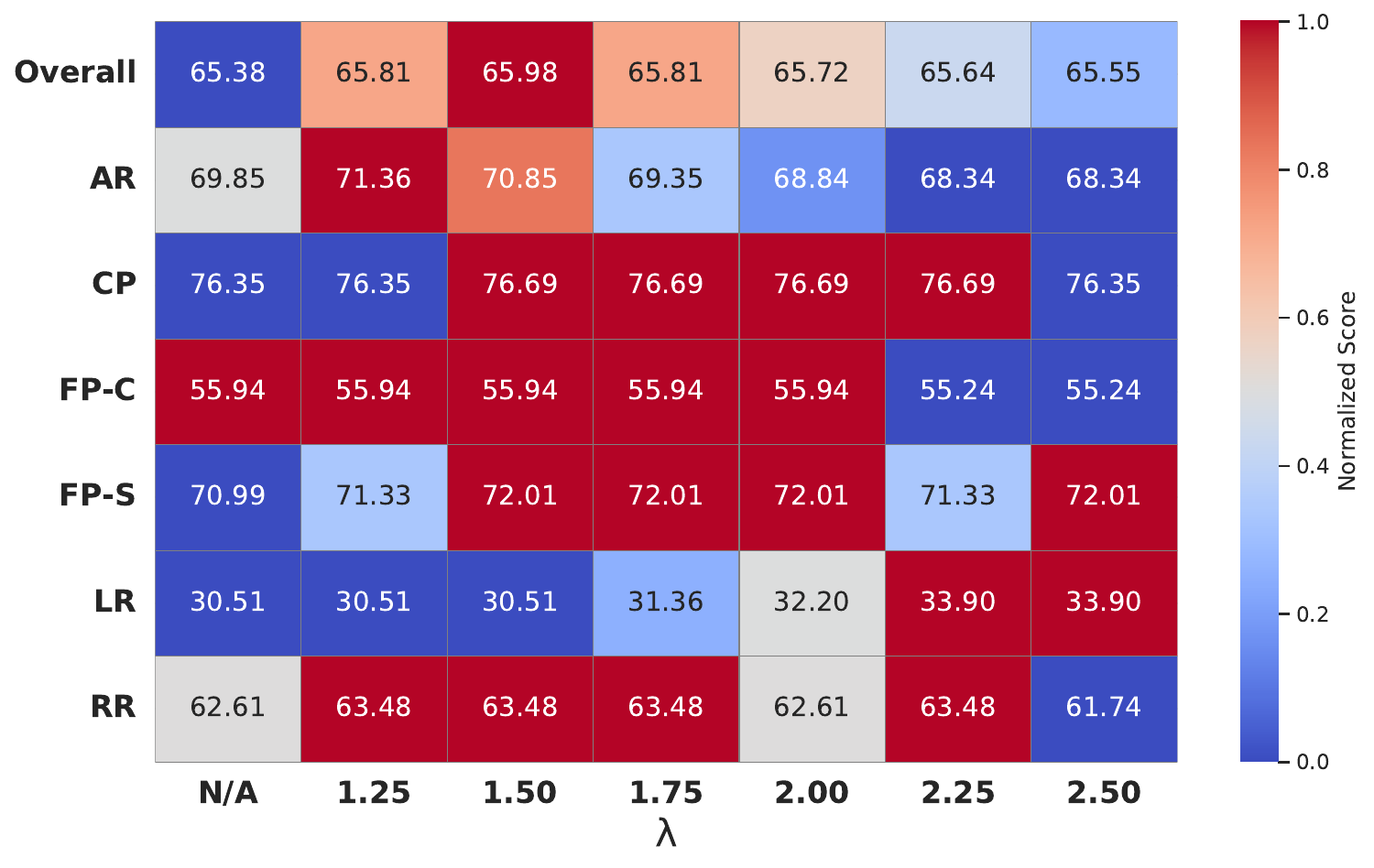}  
 \caption{Model performance on MMBench across various scaling parameter $\lambda$.}
  \label{fig:heat_mmbench}  
\end{figure}

Another essential hyperparameter is the scaling parameter $\lambda$, which is employed in soft-image guidance to regulate the guidance of the visual inputs towards the response generateion. 
Therefore, To assess the effect of varying $\lambda$ values comprehensively, we examine our method's performance on MMBench, LLaVABench and Hall-Bench with different $\lambda$ values, which can be divided into two distinct scenarios: multi-choice generation and open-end generation.
The experimental results, illustrated in \Cref{fig:heat_mmbench}, \Cref{fig:heat_llava}, and \Cref{fig:heat_hall}, suggest that an optimal value for the scaling parameter $\lambda$ lies between 1.5 and 2.0. This range provides suitable visual guidance without impairing the text generation capabilities of LVLMs.


\section{Human Evaluation on LLaVABench}
\label{sec:Human}

\begin{figure}
  \centering  
  \includegraphics[width=0.9\linewidth]{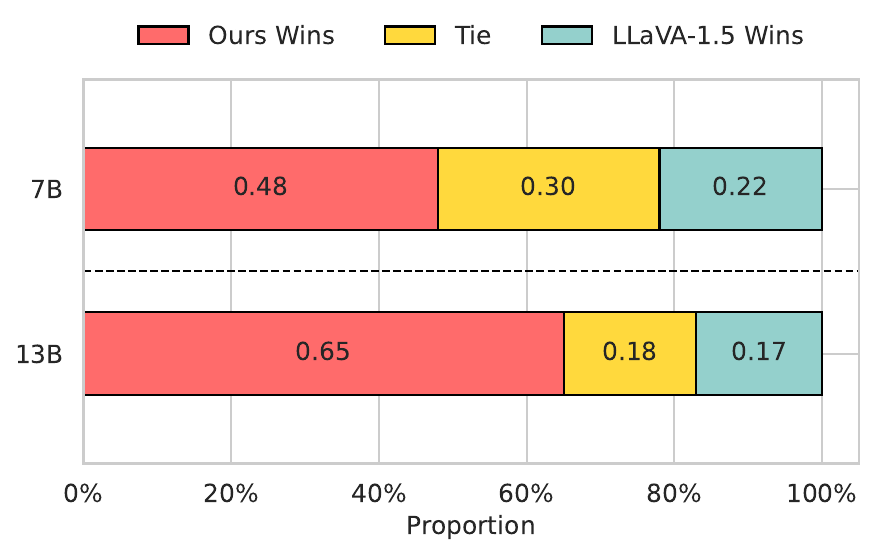}  
 \caption{Human evaluation on LLaVABench.}
  \label{fig:fig_human}  
\end{figure}

To better illustrate the efficacy of our method, a further human evaluation has been undertaken to compare the model performance of \OURS versus LLaVA-1.5~\citep{LLAVA15}. Specifically, we evaluate the model perofrmance on LLaVABench, which consists of 60 instances. We invited three human participants (all of them are Ph.D. students or Master students) to compare the responses generated by the models. For each comparison, three options were provided (Win, Tie, and Lose), with the final results determined by the majority vote of the participants.
\Cref{fig:fig_human} showcases the effectiveness of our method.

\begin{table}
  \centering  
  \scriptsize  
  \resizebox{\columnwidth}{!}{  
    \begin{tabular}{lcccc}  
      \toprule  
      $\theta$ & 5\% & 10\% & 15\% & 20\% \\
      \midrule  
      MMBench & 66.32 & \bf 66.92 & 66.75 & 65.64 \\
      LLaVABench & 67.00 & \bf 70.60  & 67.80  &  66.90 \\
      \bottomrule  
    \end{tabular}  
  }  
  \caption{ Performance of \IFG on MMBench and LLavaBench across different replace probability $\theta$ }  
  \label{tab:theta}  
\end{table}

During the human evaluation, the participants adhere the following principles to make the decision:
\begin{tcolorbox}[title={Principles of Human Evaluation for LLaVABench},  colframe=white, colbacktitle=black!65!white,breakable] 
\noindent   
You are asked to evaluate the responses generated by different models. Your evaluation should adhere to the following principles:

1.~\textbf{Correctness}: Assess whether responses address the key points outlined in the reference answer and image. For reference answers with multiple key points, evaluate how many of these the response accurately addresses and score accordingly. Additionally, ensure that the response provides the necessary information for the user. \\
2.~\textbf{Faithfulness}: Examine any additional information in the answer to verify its accuracy and relevance to the question and image. 
If this information is incorrect or not relevant to the question and image, points should be deducted. \\
3.~\textbf{Coherence}: Evaluate the fluency and coherence of the responses. Also, consider deducting points for overly verbose responses or those that are excessively generalized. \\

Finally, please make a decision among 3 opinions, including Win, Tie, and Loss.

\end{tcolorbox}

If the majority voting of three participants not yield a decisive outcome, we will engage in further discussions among the involved participants and subsequently conduct another vote to determine the final result.
The human evaluation results in \Cref{fig:fig_human} shows that \OURS can generate responses that consistently outperformed baseline models across all three evaluation criteria. These results highlight the model's ability to deliver high-quality answers that are both factually accurate and contextually relevant, while maintaining fluency and coherence.

\section{Case Studies}
\label{sec:cases}
\begin{figure}[t]  
  \centering  
  \includegraphics[width=\linewidth]{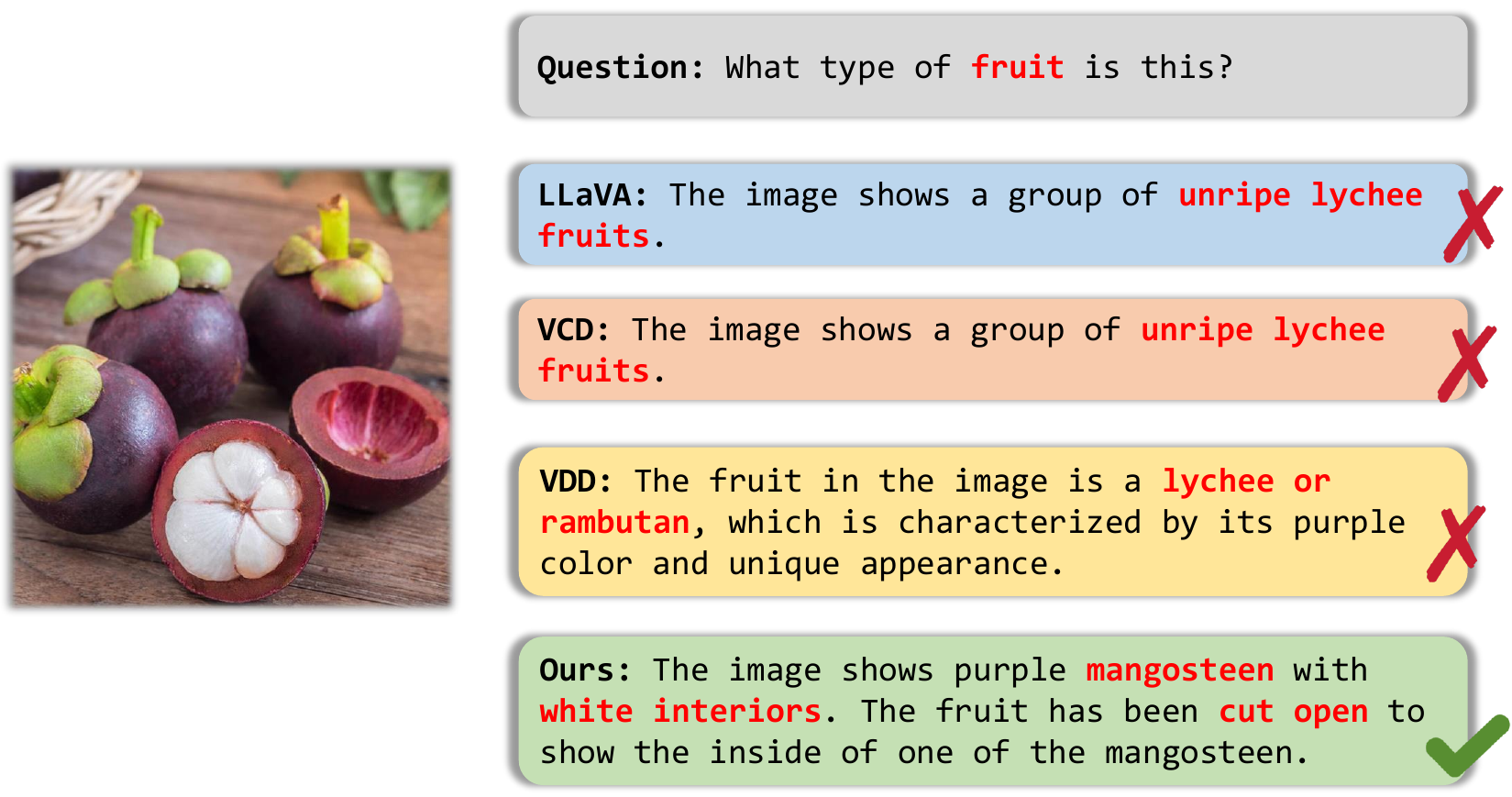}  
  \caption{Comparison of \OURS with other baselines on a sample from LLaVABench. Hallucinated responses and our corrections are highlighted in \textcolor{red}{\textbf{red}}.}  
  \label{fig:case1}  
\end{figure}

\begin{figure}[t]  
  \centering  
  \includegraphics[width=\linewidth]{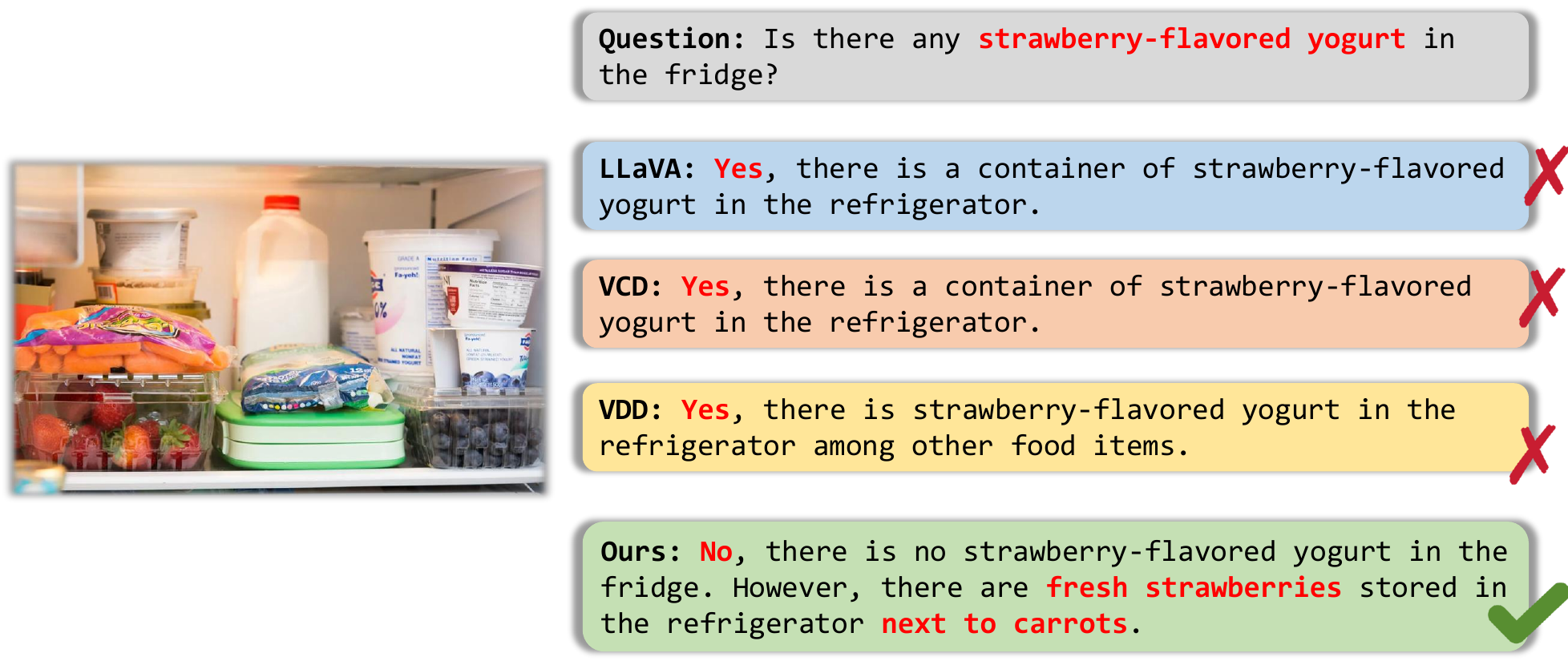}  
  \caption{Comparison of \OURS with other baselines on a sample from LLaVABench. \OURS demonstrates a reduction in object hallucination and an enhancement in fine-grained visual comprehension, such as the identification of fresh strawberries in the refrigerator.}  
  \label{fig:case2}  
\end{figure}

\begin{figure*}[t]  
  \centering  
  \includegraphics[width=\linewidth]{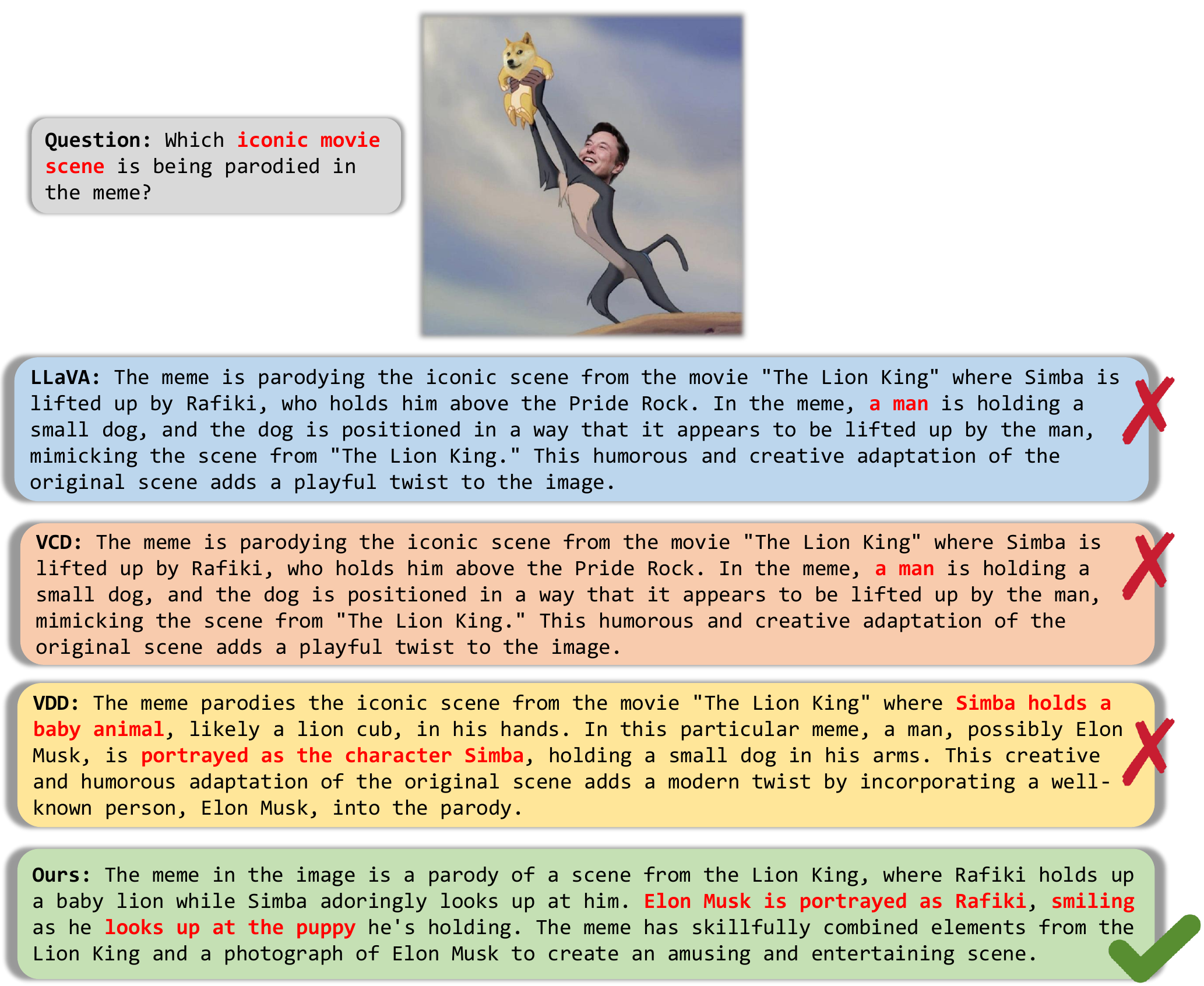}  
  \caption{Comparison of \OURS with other baselines on a sample from LLaVABench. \OURS effectively recognizes the individual in the meme and provides a thorough description, guiding to accurate responses to the question.}  
  \label{fig:case3}  
\end{figure*}

\begin{figure*}[t]  
  \centering  
  \includegraphics[width=\linewidth]{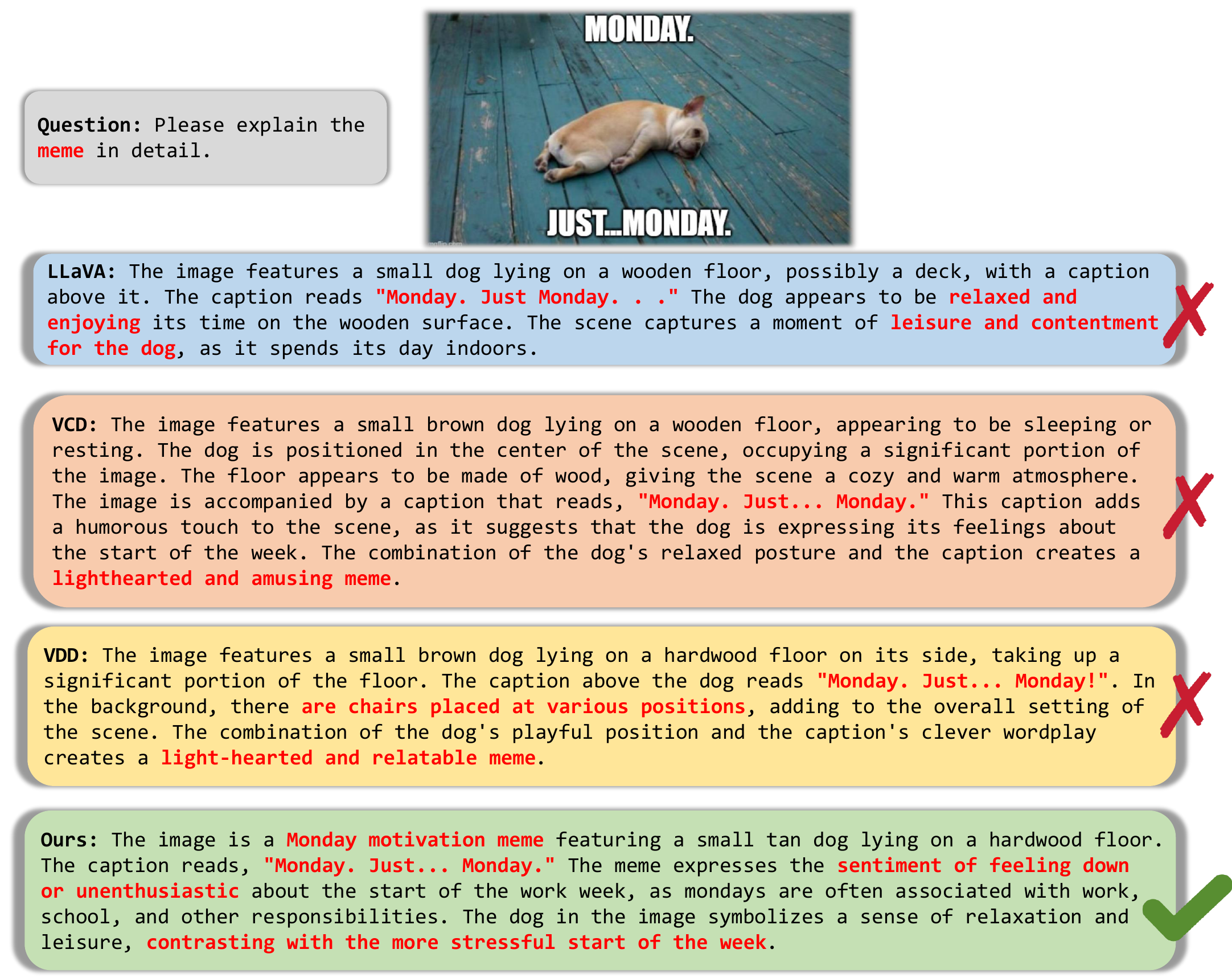}  
  \caption{Comparison of \OURS with other baselines on a sample from LLaVABench. \OURS is the only model capable of successfully articulating the idea that the meme is trying to convey, by contrasting image information and questions.}  
  \label{fig:case4}  
\end{figure*}

To deliver a thorough evaluation of the effectiveness of our methods in mitigating visual hallucinations and enhancing the visual comprehension of LVLMS, we present a case study in this section. We compare the open-ended generation results of our methods against several baseline models utilizing samples from LLaVABench. The evaluations of the case studies on the 13B model are illustrated in \Cref{fig:case1}, \Cref{fig:case2}, \Cref{fig:case3}, and \Cref{fig:case4}.
Case studies demonstrate that \OURS effectively reduces hallucinations while generating accurate responses, showcasing fine-grained visual comprehension abilities.

\section{Comparison of Attention Allocation with  Standard LVLMs}
\label{sec:visualization}
\begin{figure*}[t]  
  \centering  
  \includegraphics[width=0.70\linewidth]{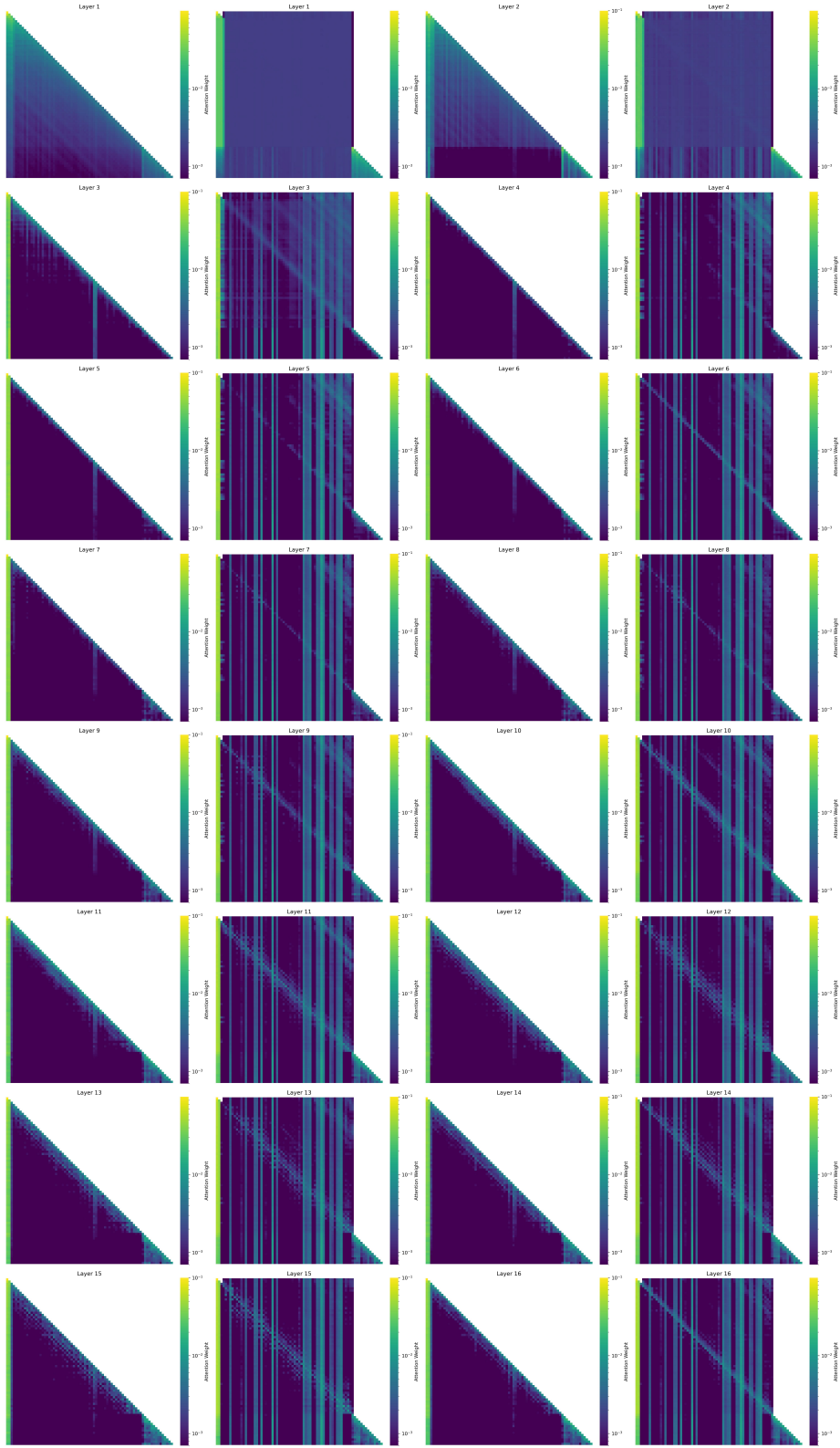}  
 \caption{Comparison of Attention Maps across the 1st to 16th Layer in LLaVA and \OURS.}
  \label{fig:full_attention_map_1}  
\end{figure*}

\begin{figure*}[t]  
  \centering  
  \includegraphics[width=0.70\linewidth]{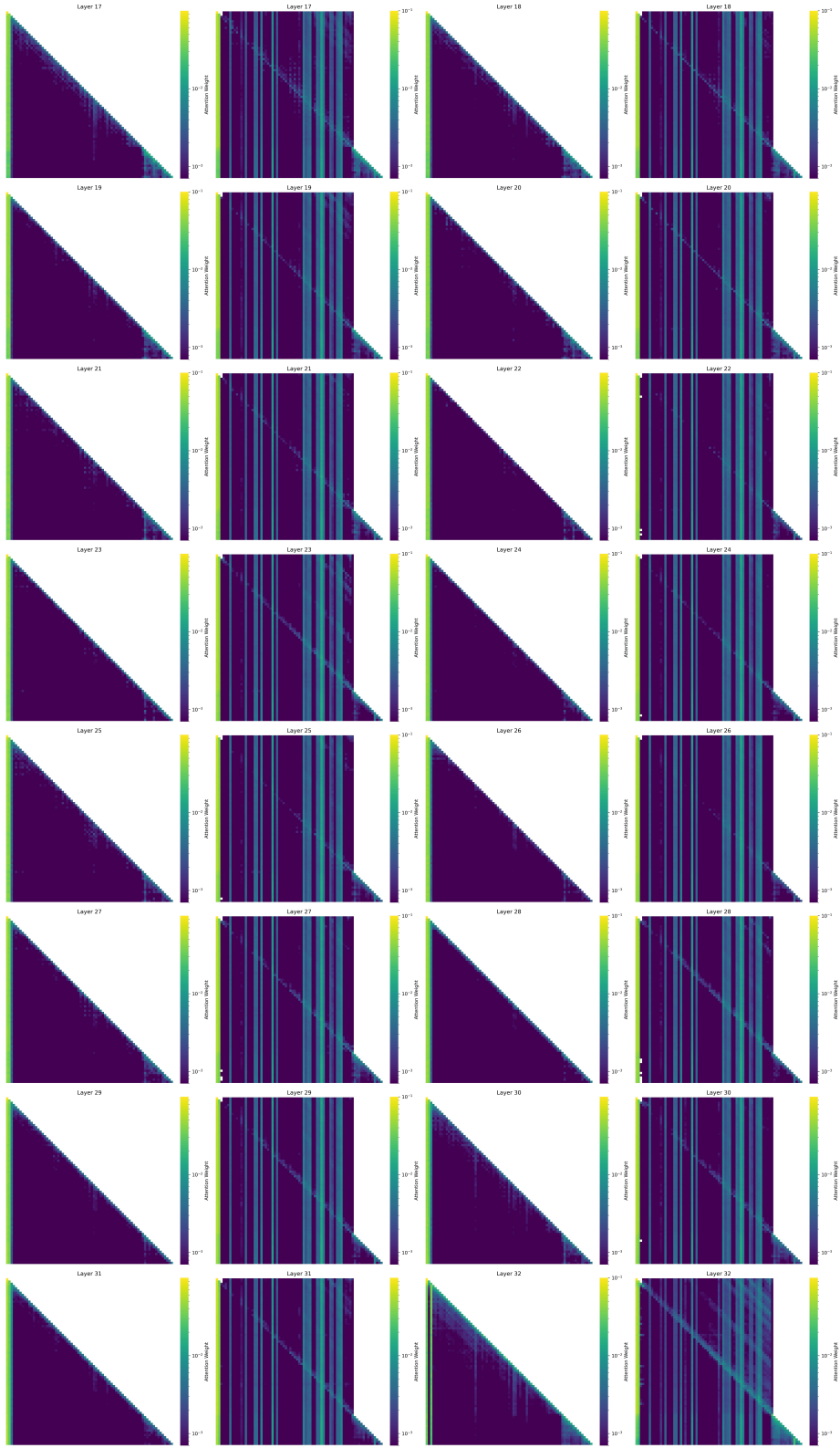}  
 \caption{Comparison of Attention Maps across the 17th to 32nd Layer in LLaVA and \OURS.}
  \label{fig:full_attention_map_2}  
\end{figure*}

To provide a clear illustration of our proposed multimodal dual-attention mechanism and distinguish the attention allocation towards visual tokens and text tokens between our model and the standard LVLM, we visualized the attention distribution across different layers in LVLMs in \Cref{fig:full_attention_map_1} and \Cref{fig:full_attention_map_2}, comparing the standard LVLM (LLaVA-1.5) with \MDA.
We get the attention allocation of LVLMs shown in this paper, such as 
\Cref{fig:full_attention_map_1}, and \Cref{fig:full_attention_map_2}, using the data from LLaVABench. We select 30 samples from the LLaVABench to query the LVLMs and examine the attention allocation during generation. 
We report the avg scores in Figure 2 and Figure 4 of the main paper.

\end{document}